\def\year{2021}\relax
\documentclass[letterpaper]{article} % DO NOT CHANGE THIS
\usepackage{aaai21}  % DO NOT CHANGE THIS
\usepackage{times}  % DO NOT CHANGE THIS
\usepackage{helvet} % DO NOT CHANGE THIS
\usepackage{courier}  % DO NOT CHANGE THIS
\usepackage[hyphens]{url}  % DO NOT CHANGE THIS
\usepackage{graphicx} % DO NOT CHANGE THIS
\def\UrlFont{\rm}  % DO NOT CHANGE THIS
\usepackage{natbib}  % DO NOT CHANGE THIS AND DO NOT ADD ANY OPTIONS TO IT
\usepackage{caption} % DO NOT CHANGE THIS AND DO NOT ADD ANY OPTIONS TO IT
\nocopyright

\usepackage{multicol}
\usepackage{comment}
\usepackage{graphicx}
\usepackage{amsmath}
\usepackage{amssymb}
\usepackage{stmaryrd}
\usepackage{paralist}
\usepackage{url}
\usepackage{hyperref}
\usepackage{cleveref}
\usepackage{lipsum}
\usepackage{gensymb}
\usepackage{breqn}
\usepackage{tabularx}

\newtheorem{definition}{Definition}

\usepackage{xcolor}
\usepackage{etoolbox}
\usepackage{caption}
\usepackage{subcaption}

\title{\LARGE \bf
Learning and Using Abstractions for Robot Planning
}

\author {
    Naman Shah,\textsuperscript{\rm 1}
    Abhyudaya Srinet, \textsuperscript{\rm 1}
    Siddharth Srivastava \textsuperscript{\rm 1} \\
}
\affiliations {
    \textsuperscript{\rm 1} Autonomous Agents and Intelligent Robots Lab, \\School of Computing, Informatics, and Decision Systems Engineering, \\ Arizona State University, \\ Tempe, AZ, USA  \\
    namanshah@asu.edu, asrinet1@asu.edu, siddharths@asu.edu 
}

\begin{document}

\maketitle
\thispagestyle{empty}
\pagestyle{empty}

%%%%%%%%%%%%%%%%%%%%%%%%%%%%%%%%%%%%%%%%%%%%%%%%%%%%%%%%%%%%%%%%%%%%%%%%%%%%%%%%
\begin{abstract}

    Robot motion planning involves computing a sequence of valid robot configurations that take the robot from its initial state to a goal state. Solving a motion planning problem optimally using analytical methods is proven to be \emph{PSPACE-Hard}. Sampling-based approaches have tried to approximate the optimal solution efficiently. Generally, sampling-based planners use uniform samplers to cover the entire state space. In this paper, we propose a deep-learning-based framework that identifies robot configurations in the environment that are important to solve the given motion planning problem. These states are used to bias the sampling distribution in order to reduce the planning time. Our approach works with a unified network and generates domain-dependent network parameters based on the environment and the robot. We evaluate our approach with \emph{Learn and Link planner} in three different settings. Results show significant improvement in motion planning times when compared with current sampling-based motion planners.
\end{abstract}

%%%%%%%%%%%%%%%%%%%%%%%%%%%%%%%%%%%%%%%%%%%%%%%%%%%%%%%%%%%%%%%%%%%%%%%%%%%%%%%%
    
\section{INTRODUCTION}

\begin{figure}[t]
    \setkeys{Gin}{width=\linewidth}
    \begin{tabularx}{\columnwidth}{X|X}
        \includegraphics{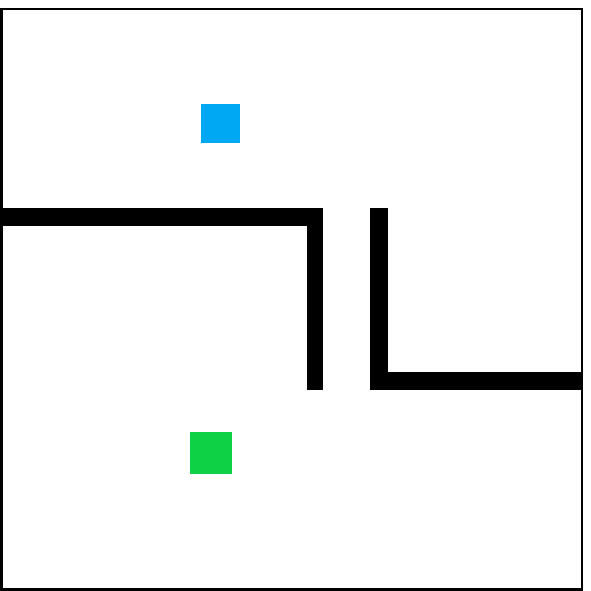} & \includegraphics{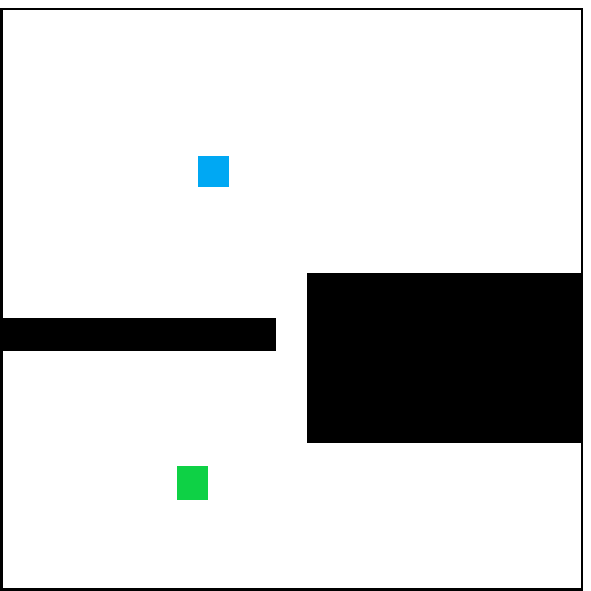} \\ 
    \end{tabularx}
    \caption{The green area represents the initial location of the robot and the
    blue area represents the goal location. Black regions show obstacles. The robot has to pass through the
    narrow passage to reach the goal location. Uniform samplers struggle in
    such scenarios due to reduced probability of locations in the narrow passage.}
    \label{fig:example_domain}
    % \vspace{-1em}
\end{figure}

Robot motion planning (MP) aims to compute a collision-free sequence of robot configurations that take the robot from an initial configuration to an intended goal configuration. The simplest motion planning problems are also proven to be \emph{PSPACE-Hard} \citep{reif1979,canny1979}. Sampling-based motion planning techniques such as rapidly-exploring random trees (RRT) \citep{lavalle1998rapidly}, probabilistic road maps (PRM) \citep{kavraki1996probabilistic}, fast matching tree (FMT) \citep{janson2013fmt}, batch informed trees (BIT*) \citep{lan2016bit} efficiently compute an approximate to the optimal solution. Such planners are known to be \emph{probabilistic complete} i.e, if the solution exists, the probability of finding a solution reaches 1.0 as the number of samples approaches infinity. Most sampling-based motion planners (SBMPs) use uniform sampling to cover the entire state space of the planning task. Such methods of sampling fail to efficiently provide samples from cluttered or narrow regions.  For example, consider the environments shown in Fig. \ref{fig:example_domain}. Green and blue regions in the images show initial and goal locations respectively. The robot has to pass through the narrow channel to reach the goal state. In such scenarios, SBMPs equipped with uniform samplers would struggle to generate samples from the narrow channel due to reduced probability. The problem becomes more challenging when the robot has a higher degree of freedom (e.g. a hinge joint or an arm manipulator) due to the dimensionality of the configuration space (\emph{C-space}).

\begin{figure*}[t]
 \vspace{0.3em}
    \begin{center}
        \includegraphics[width=\textwidth]{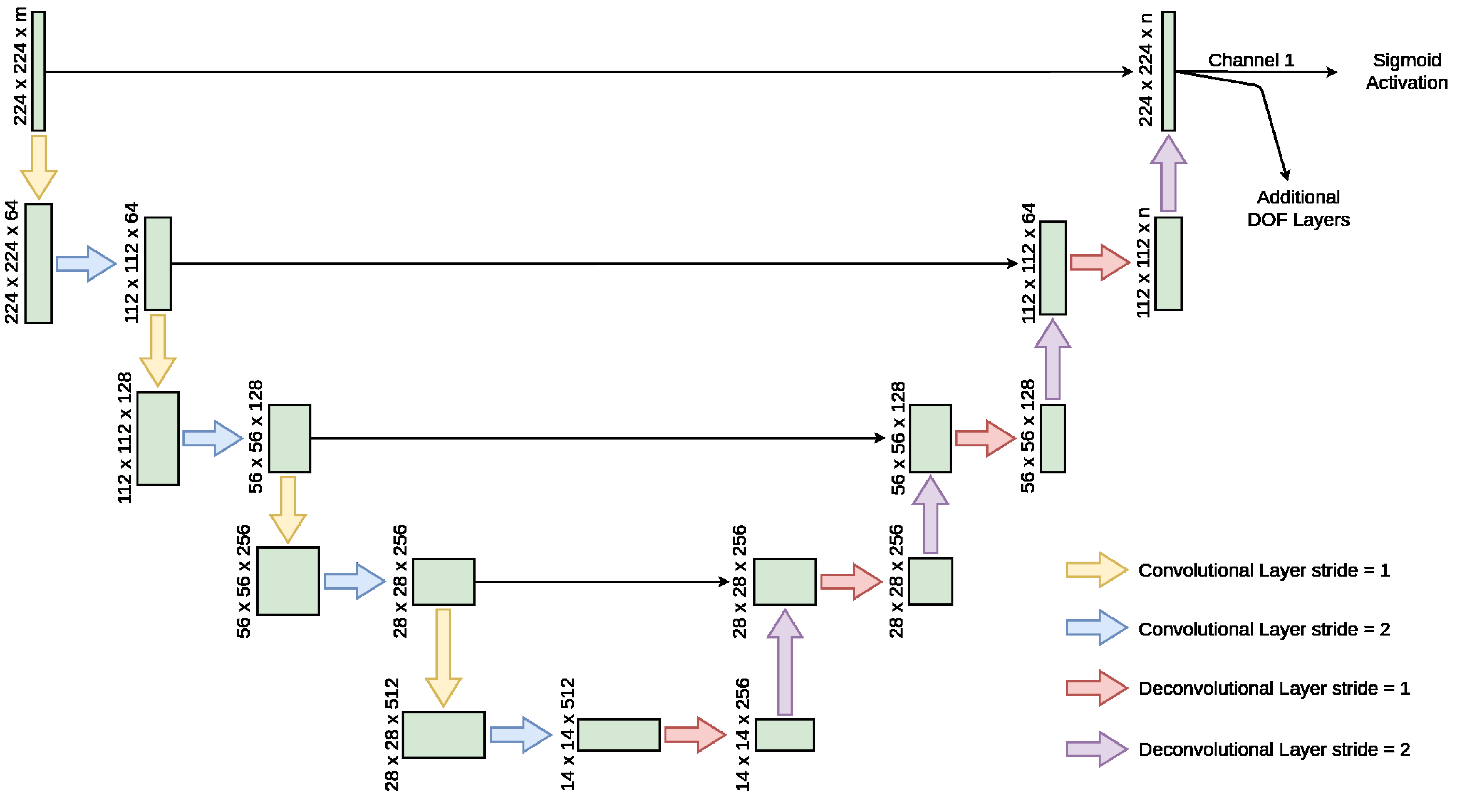}
   
    \caption{Network Architecture. Number of input channels $m$ and number of output channels $n$ depends on the environment and the robot.}
    \label{fig:nn_arch}
    \end{center}
    % \vspace{-1.5em}
\end{figure*}

This paper focuses on speeding up sampling-based motion planning by identifying regions that are important to solve the given problem. These regions (sets of robot configurations) or configurations are the ones that the robot must achieve to solve the posed motion planning problem. Such regions are analogous to the concept of \emph{landmarks} in automated task planning.  \citet{dan_llp} defines such sets as \emph{critical regions} and use them to improve efficiency of path planning problems in \emph{2D spaces}. Such critical regions can also be used to facilitate a hierarchical algorithm that first computes a path over the identified critical regions and use SBMP over these regions. In this paper, we propose a deep-learning-based framework that identifies such critical regions for any given arbitrary motion planning problem and uses them to bias the sampling distribution for an SBMP. With this biased distribution, the sampling-based motion planning algorithms are more likely to sample states from these critical regions and that should lead to reduced planning times.

In this paper, the proposed approach uses demonstrations to learn \emph{critical regions} for the given motion planning task. Our approach utilizes a unified network. It uses the \emph{C-space} to compute network architecture and domain-dependent parameters of the network and utilize it to learn the distribution of critical regions for the given \emph{C-space}. We evaluate our approach in three different domains with three different robots.  Our empirical evaluation shows that the model is able to capture important information such as environment and robot geometry, critical states, and goal states. Results show significantly reduced planning times compared to recently developed SBMPs that use uniform sampling. 

To summarize the main contribution of our approach, our approach is able to use the number of degrees of freedom of the robot and the dimensionality of \emph{C-space} to generate a robot-specific network architecture to learn the distribution of critical configurations. It uses the learned model to predict critical regions for the new motion planning problem, which is then used with a sampling-based motion planner to efficiently compute a motion plan. 

The paper is organized as follows: section \ref{sec:related} reviews recent work that addresses similar problems. Section \ref{sec:method} provides the detailed understanding of our approach, section \ref{sec:evaluation} discusses the empirical evaluation of our approach and compares it with recent motion planning algorithms, and section \ref{sec:ablation} discusses various approaches tried before the presented approach.

\section{RELATED WORK}
\label{sec:related}

There has been a lot of research on guiding SBMPs to improve the speed and efficiency of motion planning. Heuristically guided  RRT~\citep{heuristic_rrt} efficiently uses heuristics to guide the search towards the goal. Though this approach is useful to compute motion plans with optimal cost, such an approach would be able to work only after sufficient samples are generated to cover the entire configuration space. \citet{Burns2005TowardOC} propose to use an adaptive sampling approach that estimates the model of the \emph{C-space} and its connectivity. \citet{Havoutis2009} learn a non-linear manifold of the \emph{C-space} to estimate accurate heuristic of the states in the \emph{C-space}. \citet{yang2004} use decomposition techniques to estimate the medial axis of the state-space model to bias the sampling distribution. While the heuristic obtained using the medial axis may be effective, it does not provide any guarantees of increased performance. ~\citet{zhang2018learning} use rejection sampling to reject unrelated samples to speed up the motion planning problem. ~\citet{zhang2018learning} use reinforcement learning to learn a policy that decides to accept or reject a new sample to expand the search tree. While it reduces the search space to compute the path, it may suffer from higher sample rejections in environments that have narrow passages which are not easily sampled by uniform samplers. While all of these approaches try to estimate the heuristic for improving the quality of motion plans and reducing planning time, they still require an efficient sampler to provide the samples. Our approach aims to learn such a sampler that can be used to efficiently generate samples that are required to solve the problem.

 \begin{figure}[t]
 \vspace{0.5em}
 \setkeys{Gin}{width=\linewidth}
    \begin{tabularx}{\columnwidth}{X|X|X}
        \includegraphics{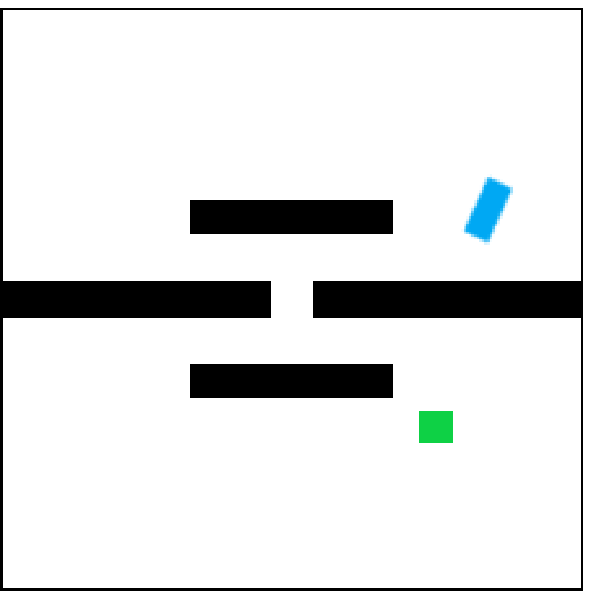} &  \includegraphics{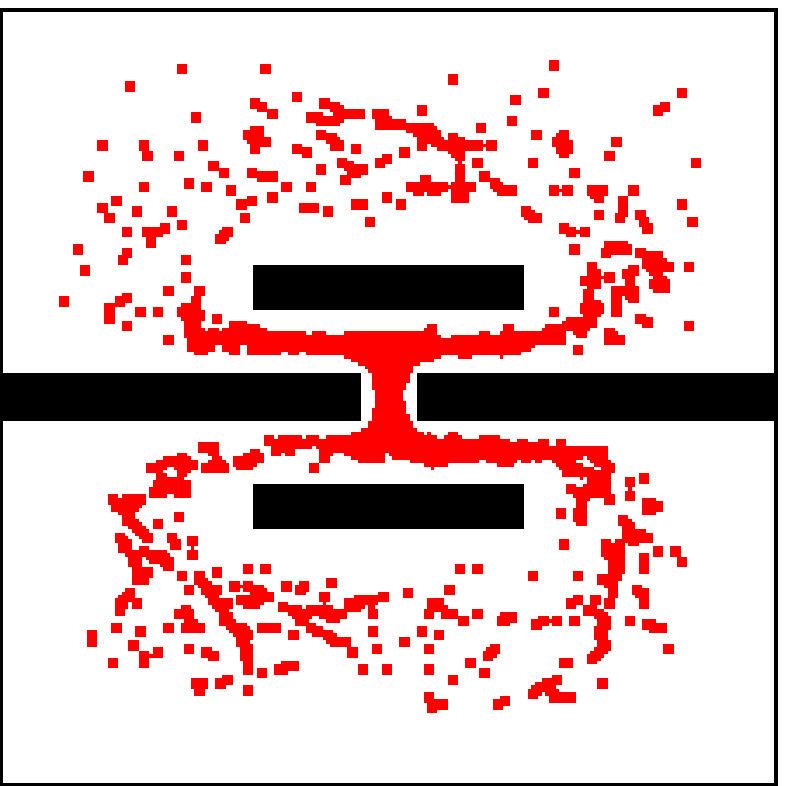} &  \includegraphics{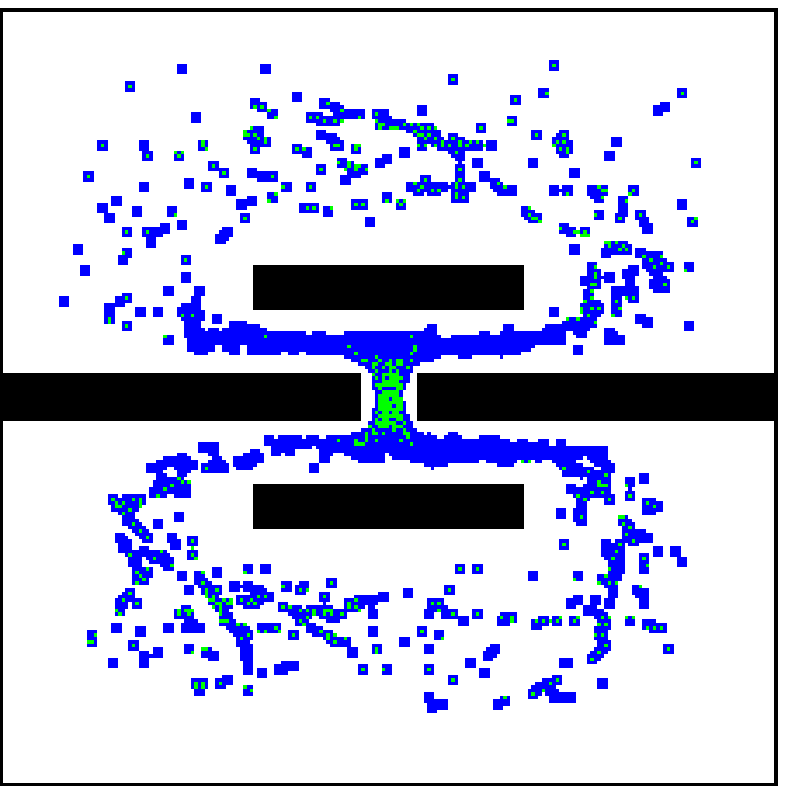} \\
        \includegraphics{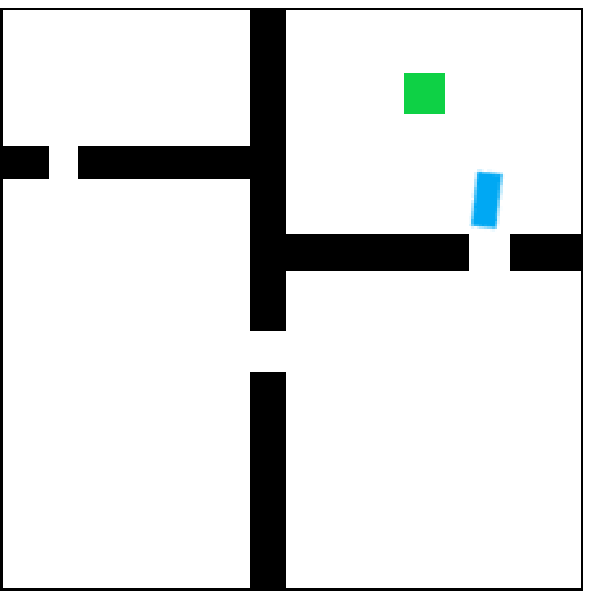} &  \includegraphics{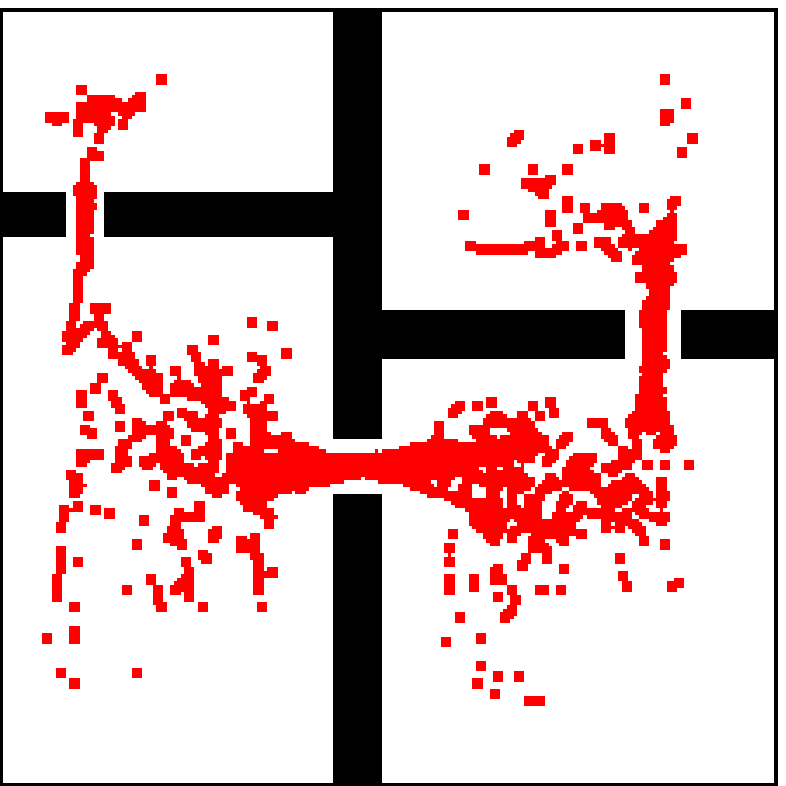} &  \includegraphics{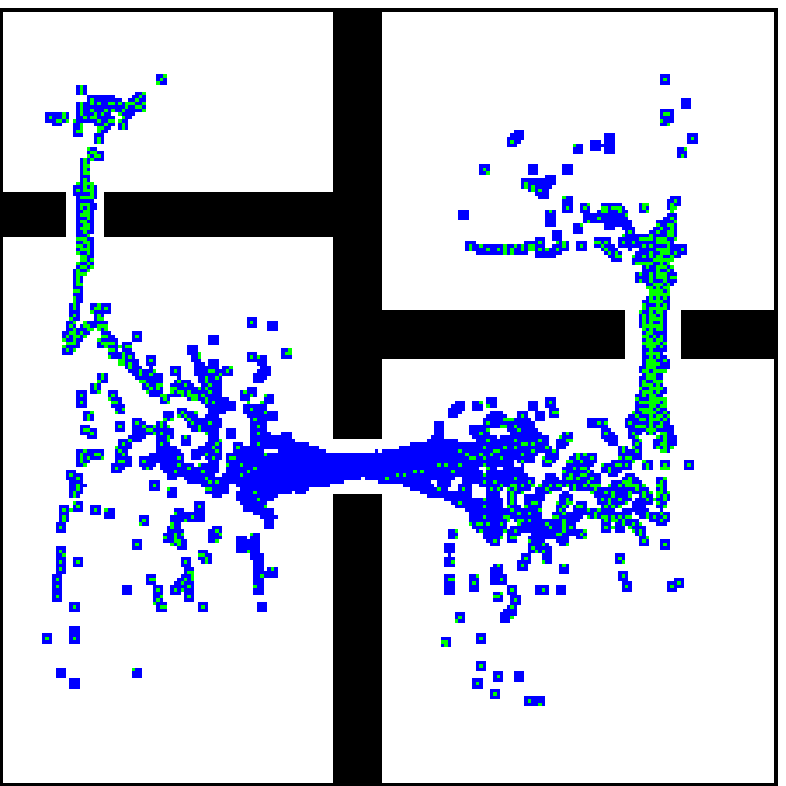} \\
        \centering (a) & \centering (b) & \centering (c) \\
    \end{tabularx}
    \caption{Training environments and labels for $\mathcal{S}\mathcal{E}^2$ domain. (a) represents the input to the environment. Each label is generated by solving $50$ motion planning problems. Red regions in (b) are critical locations of environments that are important to solve the problem. Critical regions are computed using Def. \ref{def:critical}. (c) shows the distribution for the orientation of the robot. Blue regions show horizontal alignment of the robot and the green regions show vertical alignment of the robot. Blue rectangle in (a) shows the robot.}
    \label{fig:trining_envs_3}
    % \vspace{-1em}
\end{figure}

Multiple approaches have tried using statistical learning to boost motion planning. ~\citet{wang2021survey} present a comprehensive survey on methods that utilize a variety of learning methods to improve the efficiency of the SBMPs. We discuss a few of these approaches that are relevant to our approach. ~\citet{qureshi2020motion} use a convolutional neural network model using the data obtained from video streaming and expert demonstrations to guide the search exploration for path planning. ~\citet{ichter2018learning} and ~\citet{kumar2019} use a conditional variational autoencoder (CVAE) \citep{Sohn2015LearningSO} to learn the sampling distribution for the motion planning problem. Compared to the generative model presented by \citet{ichter2018learning}, our discriminative model requires less amount of data and provides a stable learning framework that requires less complex input representation. The simpler representation allows our approach to scaling to higher dimensional robots and reduces overall inference time. ~\citet{dan_llp} use an image-based approach to learn and infer the sampling distribution using demonstrations. They use top view images of the environment with \emph{critical regions} highlighted in the image to learn the distribution of critical regions. While their approach efficiently learns critical regions for simple navigational problems, the representation used for the labels restricts their approach to such simple navigational problems and doesn't allow them to work with arbitrary motion planning problems that have \emph{C-spaces} of dimensions higher than two. ~\citet{liu2020icra} use semantic information to bias the sampling distribution for path planning tasks in partially known environments. Compared to \citet{liu2020icra}, our approach is not limited to path planning problems and does not require semantic information explicitly but aims to learn such a notion in the form of critical regions.

 \begin{figure}[t]
  \vspace{0.5em}
    \setkeys{Gin}{width=\linewidth}
    \begin{tabularx}{\columnwidth}{X|X|X|X}
        \includegraphics{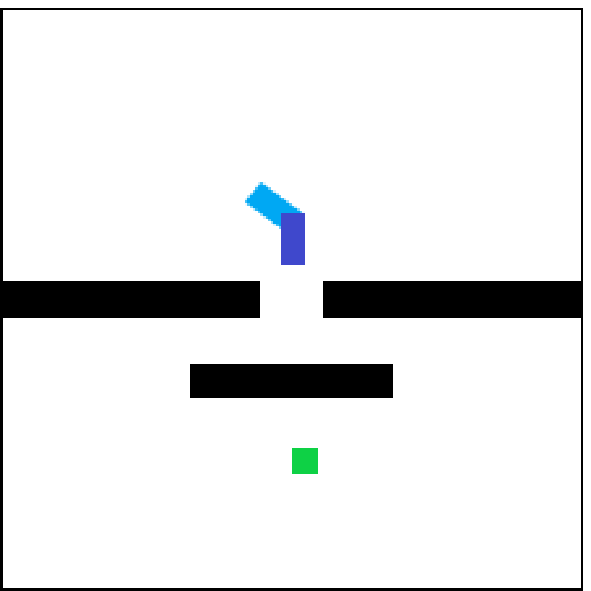} &  \includegraphics{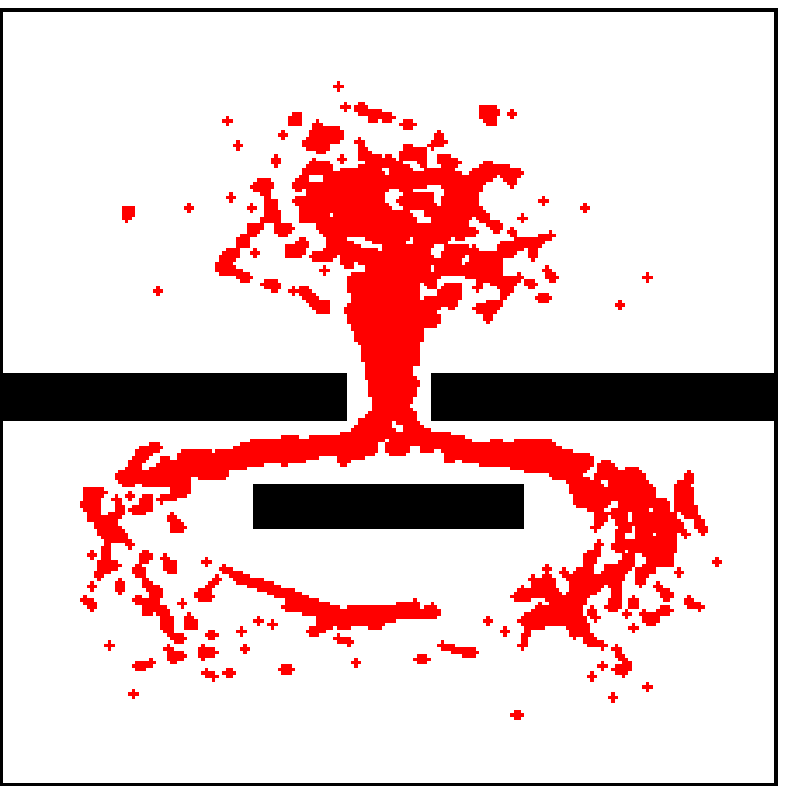} &  \includegraphics{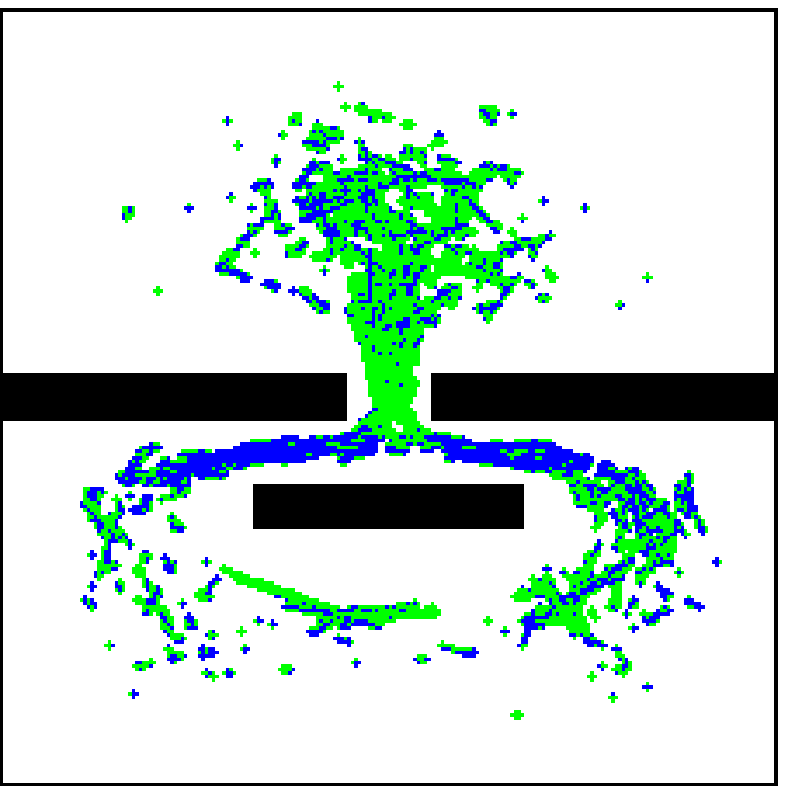} & \includegraphics{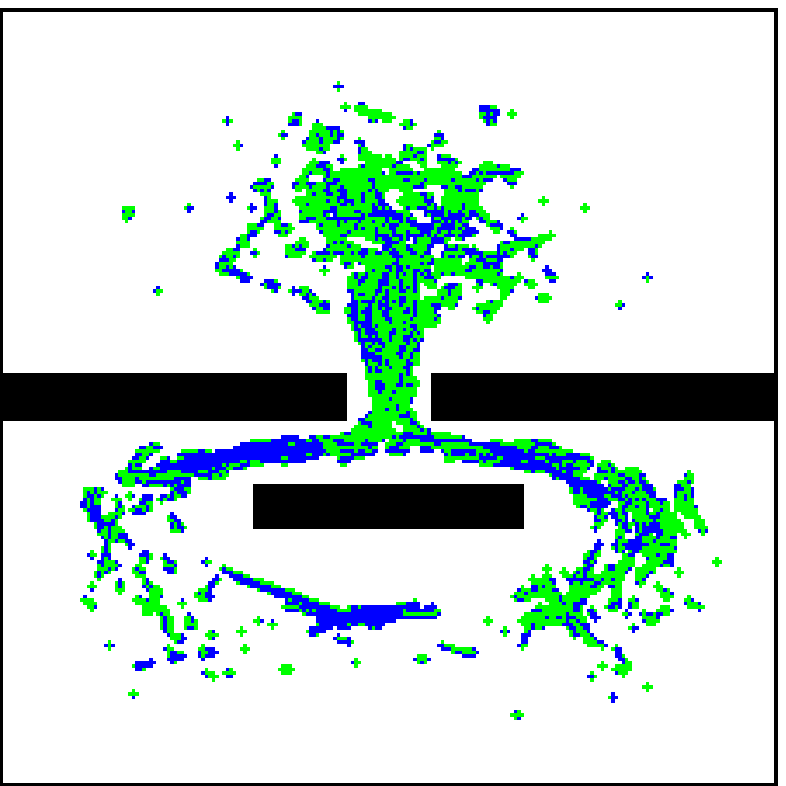} \\
        \includegraphics{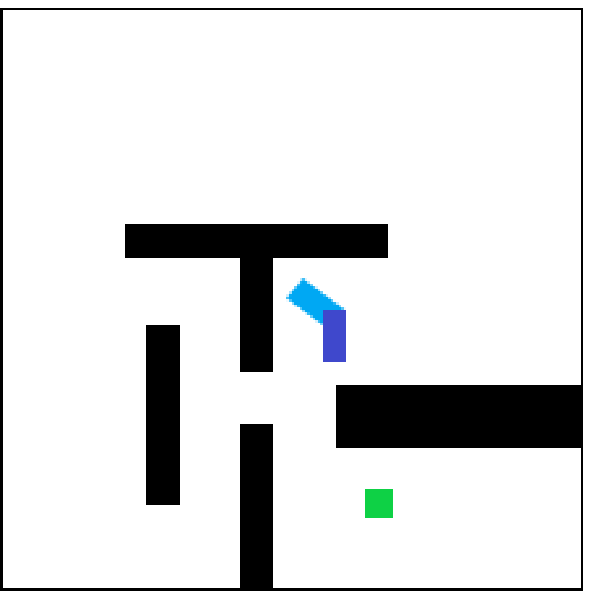} &  \includegraphics{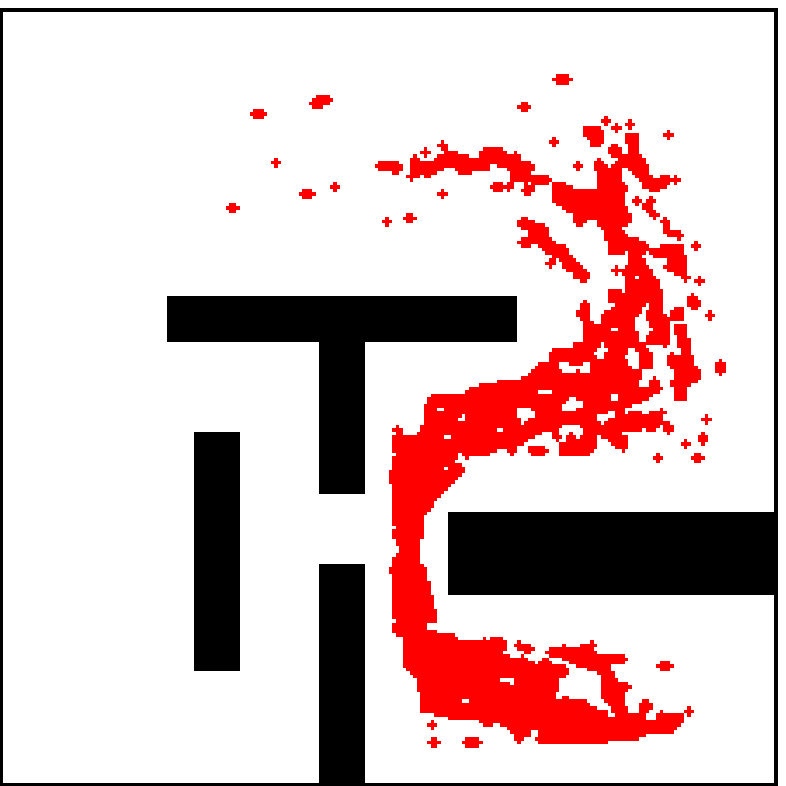} &  \includegraphics{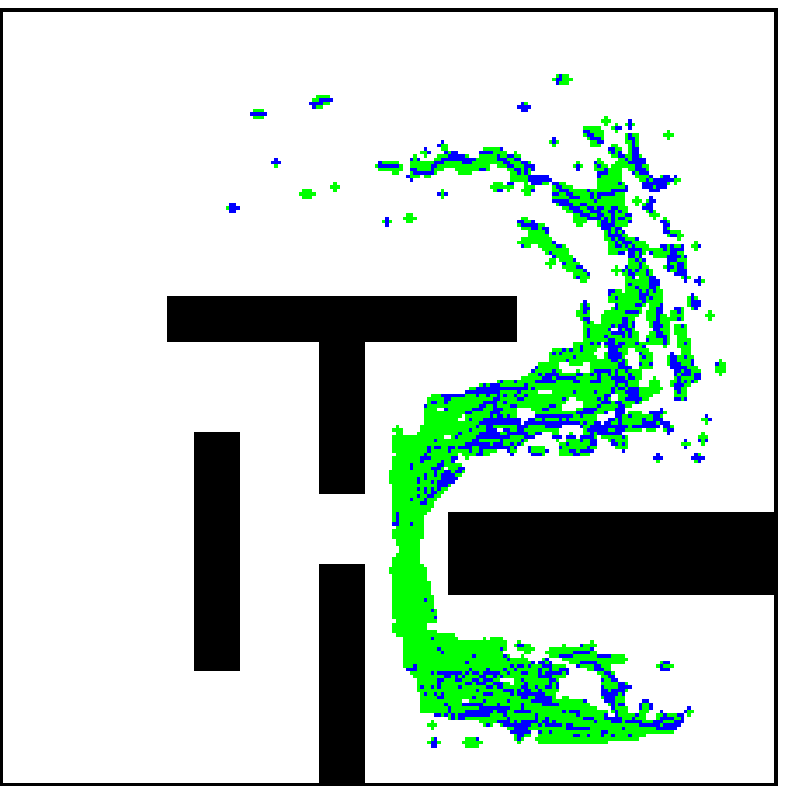} &  \includegraphics{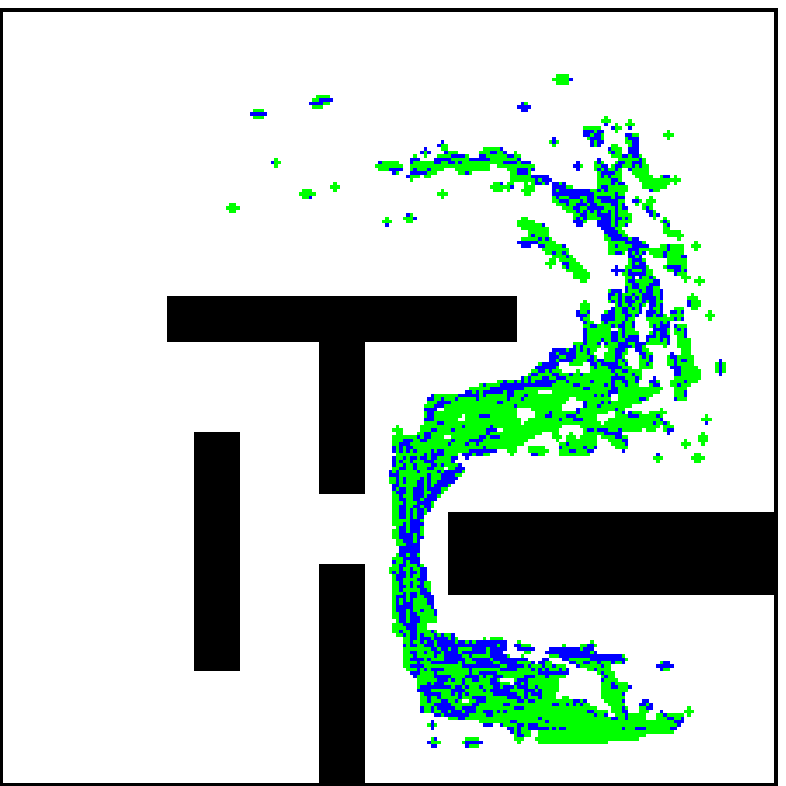}\\
        \centering (a) & \centering (b) & \centering (c) & \centering (d) \\
    \end{tabularx}
    \caption{Training environments for \emph{hinged robot} domain. (a) represents the input to the environment. Each label is generated by solving $50$ motion planning problem. Red regions in (b) are critical locations of the environment. (c) shows the distribution of the orientation for the base link of the robot. Blue regions show the horizontal and green regions show the vertical alignment of the robot. (d) shows the distribution for the joint $\omega$. Blue regions show that $\omega$ is closer to $180^{\circ}$ and green regions show that $\omega$ is close to either $90^{\circ}$ or $270^{\circ}$.Blue rectangles in (a) show the robot with the hinged joint.}
    \label{fig:training_envs_4}
    % \vspace{-1em}
\end{figure}

\section{METHODOLOGY}
\label{sec:method}

We use the notion of \emph{critical regions} in an environment as defined by \citet{dan_llp}:
\begin{definition}
    \label{def:critical}
    Given a robot $R$, an environment $E$, and a class of motion planning problems $M$, the measure of criticality of a Lebesgue-measurable open set $r \subseteq \mathbb{C}$, $\mu(r)$, is defined as $\lim_{s_n \to ^{+}r} \frac{f(r)}{v(s_n)} $, where $f(s_n)$ is the fraction of observed motion plans solving tasks from $M$ that pass through $s_n$, $v(s_n)$ is the measure of $s_n$ under a reference density, and $\to^{+}$ denotes the limit from above along any sequence $\{s_n\}$ of sets containing $r$ ($r \subseteq s_n$, $\forall\,n$). 
 \end{definition}

 The above definition identifies regions that are important to solve a path planning problem. It assigns a higher score to the regions in the environment that have a higher fraction of motion plans passing through it. We use this notion of \emph{critical regions} to compute sets of critical configuration for the given robot using motion planning demonstrations. 
%  \begin{definition}
%      Given a robot $R$, and environment $E$, and a class of motion planning problems $M$, the measure of criticality of a robot state or a configuration $c$, $\mu(c)$, is defined as $\frac{f(c)}{u}$, where $f(c)$ is the fraction of motion plans solving tasks $M$ that has configuration $c$ as their intermediate waypoint and $u = max_{c} f(c)$. A state or a configuration $c$ is identified as a \emph{critical state} or \emph{critical configuration} if the measure of criticality $\mu(c)$ is greater than or equal to some threshold $t$. 
%  \end{definition}

 \subsection{Learning Framework}

We now explain the overall learning framework. We start by explaining how our network is able to generate a robot-specific architecture of the network and then discuss the data generation and network training. 
 \subsubsection{Deriving Robot Specific Network Architectures} ~\\
 In this section, we explain how our approach is able to generate robot-specific network architectures using the number of degrees of freedom and the robot geometry. We use these robot specific information to update the UNet \cite{Ronneberger2015unet} base architecture as shown in the Fig. \ref{fig:nn_arch}. We start with a fully convolutional UNet architecture that consists of a total of sixteen layers with $3\times3$ filters. The first eight layers are convolutional layers that work as an encoder unit and the remaining eight layers are deconvolutional layers that work as a decoder unit.  We use a single channel of the last base layer to predict critical locations for the end-effector (base link for path planning) in the environment. We use additional $k$ convolutional layers to predict the distribution for joint values. Here $k$ represents the number of degrees of freedom the robot that is not captured by the location of the end-effector in the environment. For e.g., for a robot that has four degrees of freedom $(x,y,\theta,\omega)$ where $x$ and $y$ represents the location of the robot in the environment, the network would have $k=2$ additional convolutional layers - one convolutional layer for each remaining degree of freedom ($\theta$ and $\omega$ in this case). The dimensions of input to the network are computed using the dimensionality of the environment and the number of degrees of freedom of the robot. The first channel in the input represents the occupancy matrix of the environment and each additional channel represents the goal value for each degree of freedom of the robot. So for a robot with four degrees of freedom, the network will have five channels in the input layer.

 \begin{figure}[t]
 \vspace{0.3em}
    \setkeys{Gin}{width=\linewidth}
    \begin{tabularx}{\columnwidth}{X|X|X}
        \includegraphics{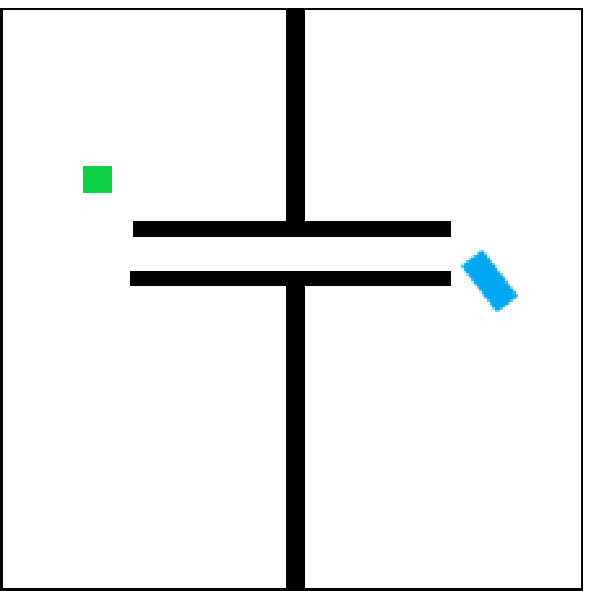} &  \includegraphics{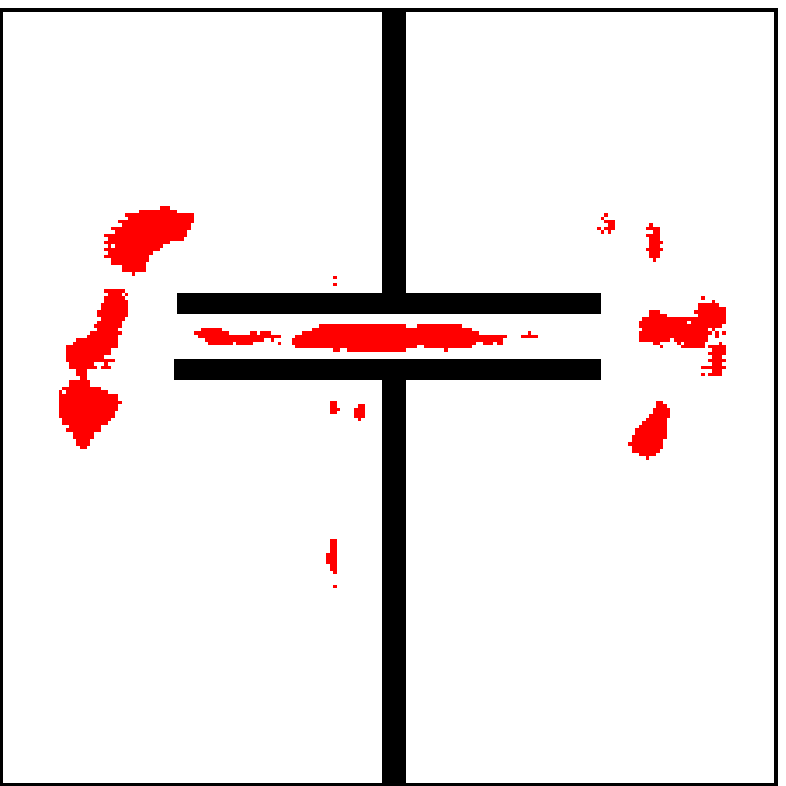} &  \includegraphics{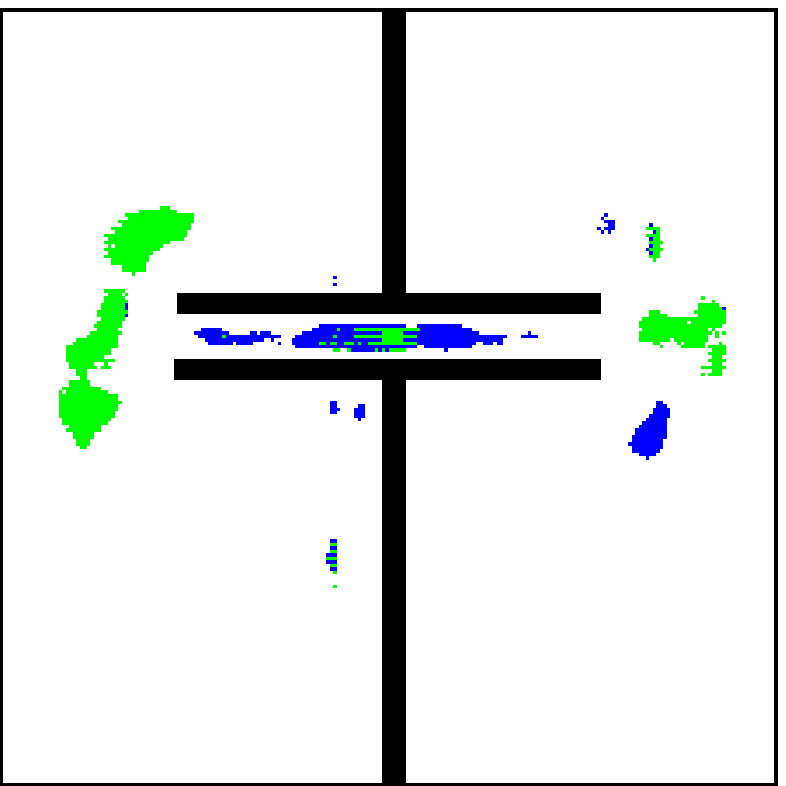} \\
        \includegraphics{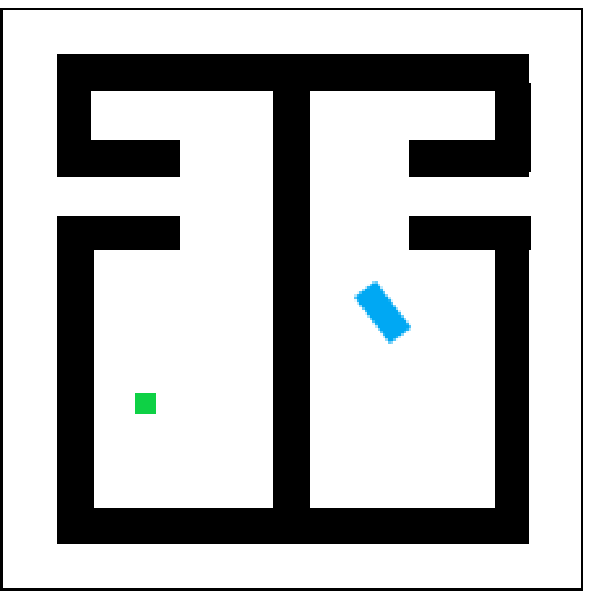} &  \includegraphics{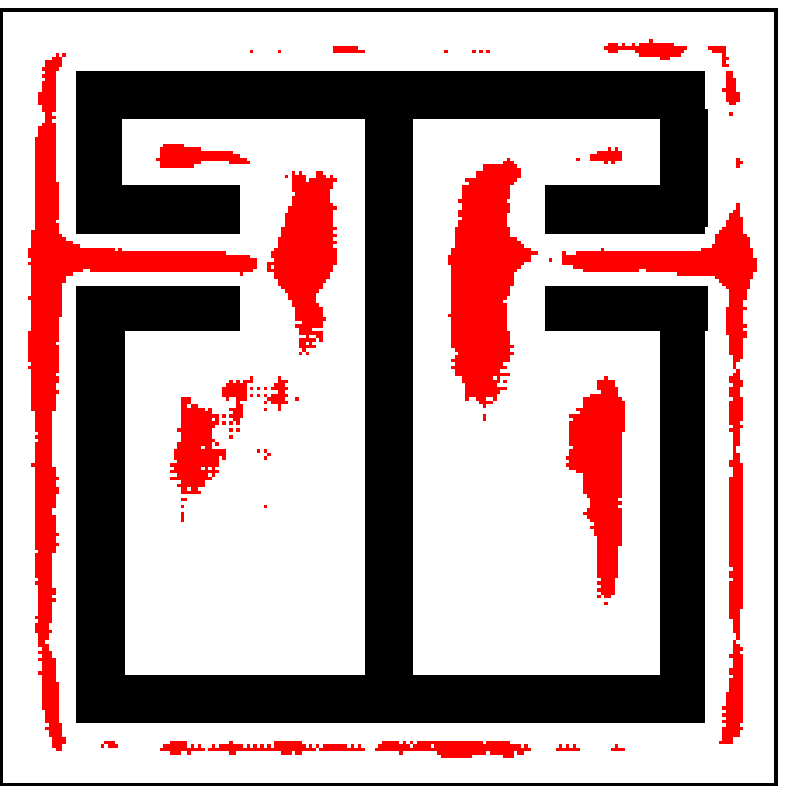} &  \includegraphics{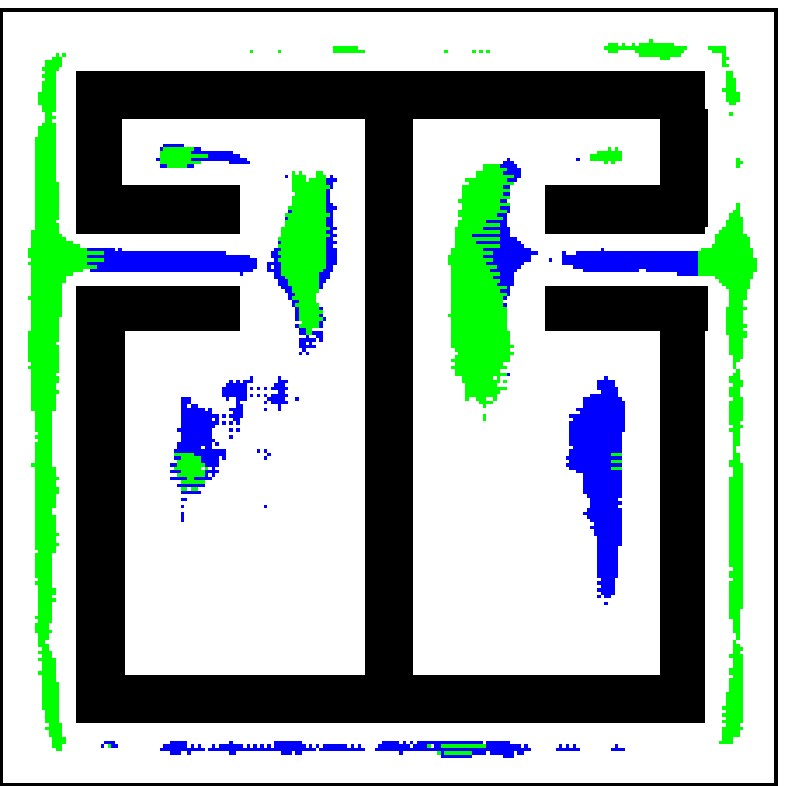} \\
        \centering (a) & \centering (b) & \centering (c) \\
    \end{tabularx}
    \caption{Test evaluations for $\mathcal{S}\mathcal{E}^2$ domain. (a) represents the input to the environment. Top: Environment 3A, bottom: Environment 3B. Red regions in (b) are locations of environments that are predicted critical by the network. (c) shows the predicted orientation of the robot. Blue regions are locations where the network predicted robot to be horizontal and green regions are the regions where the network predicted the robot to be vertical.}
    \label{fig:test_3dof}
    % \vspace{-1em}
\end{figure}

 \begin{figure}[t]

 \vspace{0.3em}
    \setkeys{Gin}{width=\linewidth}
    \begin{tabularx}{\columnwidth}{X|X|X|X}
        \includegraphics{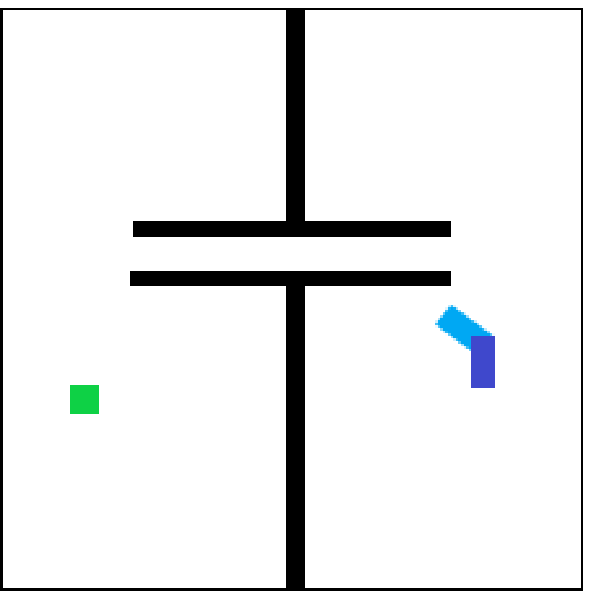} &  \includegraphics{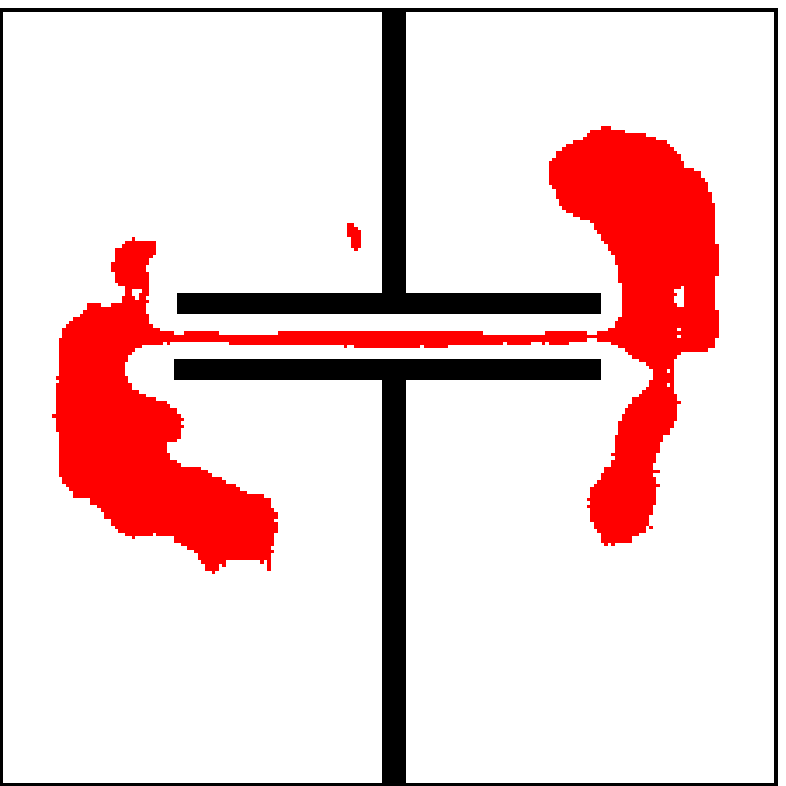} &  \includegraphics{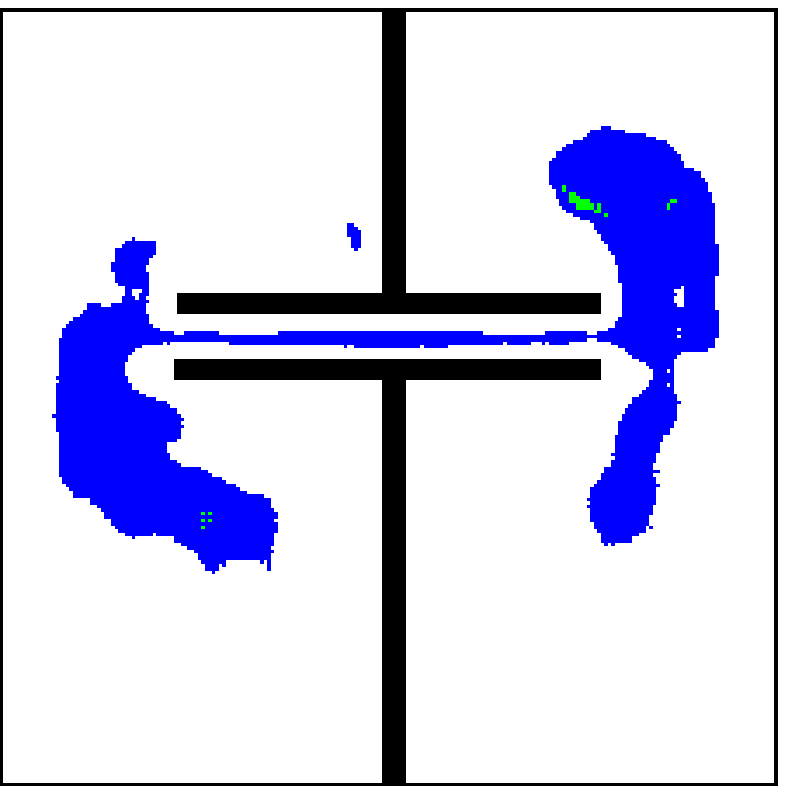} & \includegraphics{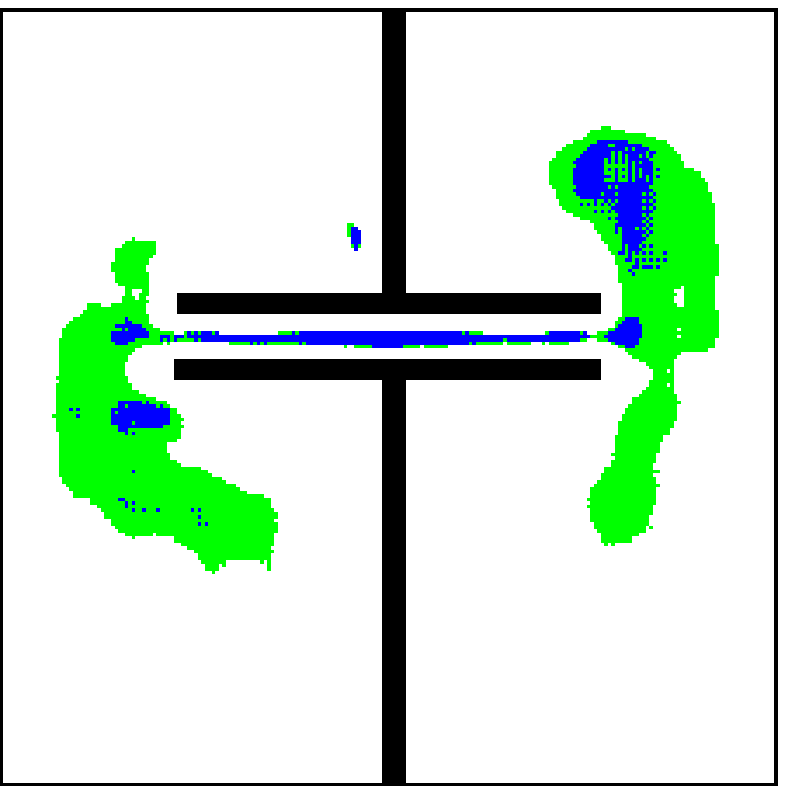} \\
        \includegraphics{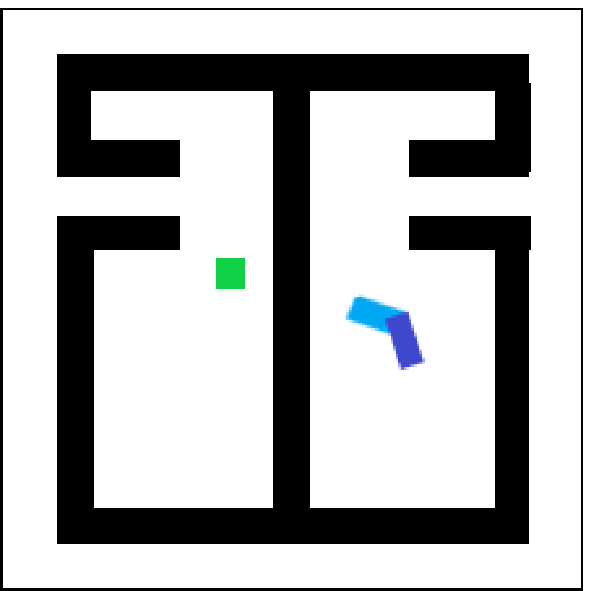} &  \includegraphics{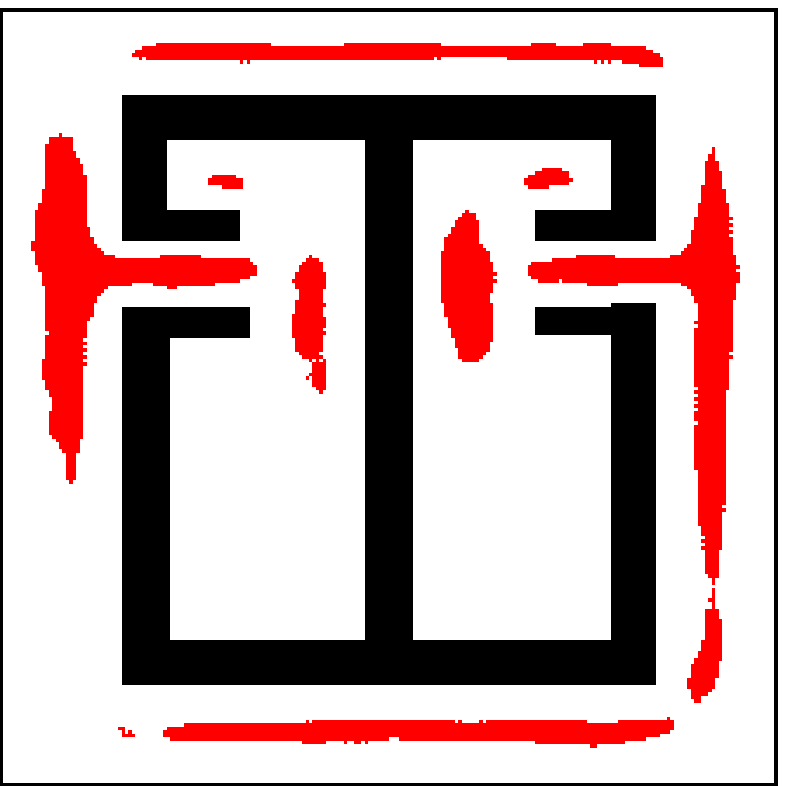} &  \includegraphics{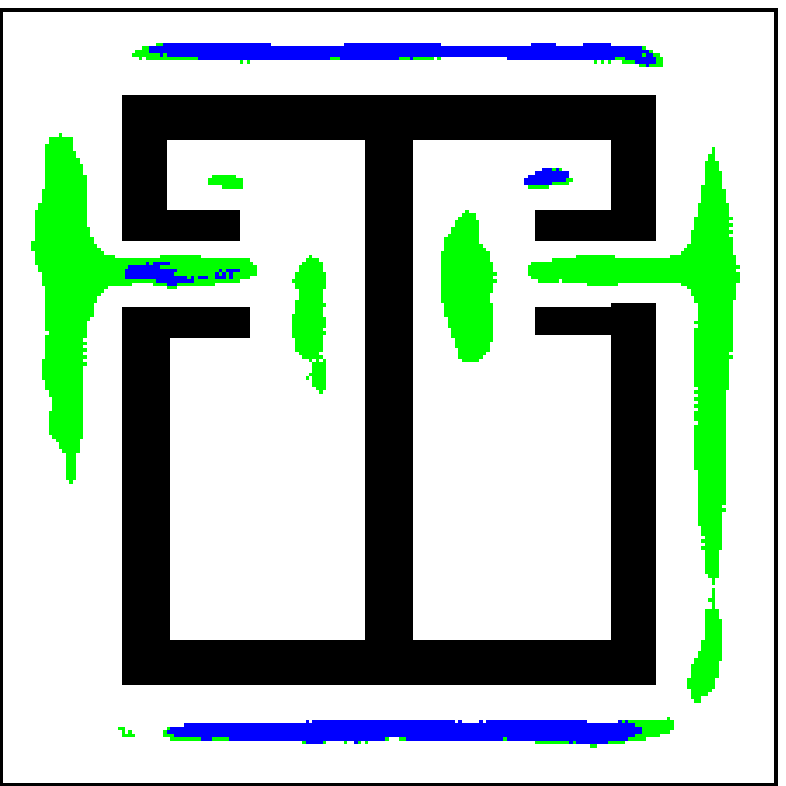} &  \includegraphics{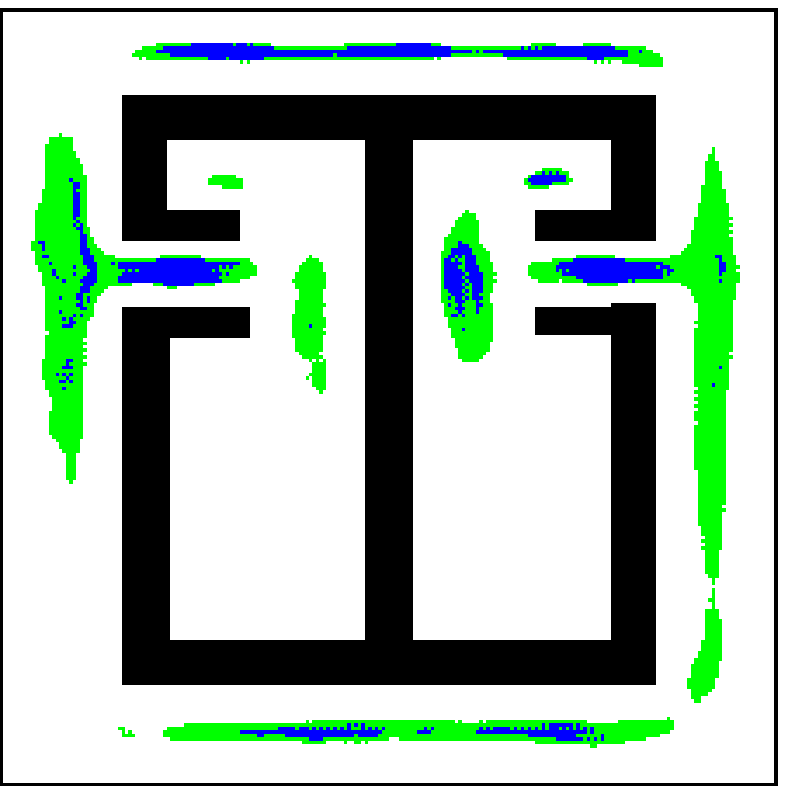}\\
        \centering (a) & \centering (b) & \centering (c) & \centering (d) \\
    \end{tabularx}
    \caption{Test evaluations for \emph{hinged robot} domain. (a) represents the input to the environment. Top: Environment 4A, bottom: Environment 4B. Red regions in (b) are locations of environments that are predicted critical by the network. (c) shows the predicted orientation for the base link of the robot. Blue regions are locations where the network predicted the base link to be horizontal and green regions are the regions where the network predicted the base link to be vertical. (d) shows the distribution for the joint $\omega$. Blue regions show that the network predicted it to be closer to $180^{\circ}$ and green regions show that the network predicted the angle $\omega$ close to $90^{\circ}$ or $270^{\circ}$.}
    \label{fig:test_4dof}
    % \vspace{-1em}
\end{figure}

  The total number of channels in a label vector is also computed using the number of joints in the robot. We discretize the values of each joint of the robot in $p$ discrete bins. We use a single channel to represent the distribution of critical regions for the end-effector.  Additional $p$ channels represent the distribution for values of each joint. So for a robot with four degrees of freedom $(x,y,\theta,\omega)$, where $x$ and $y$ represents the location of the robot in the environment, the total number of output channels would be equal to $21$, if $p=10$ for $\theta$ and $\omega$.

 \subsubsection{Training Data Generation}
 \label{subsec:data}
For each instance of an environment, we generate a set of $100$ motion planning tasks with randomly generated goal states. For each motion planning task, we generate $50$ sub-tasks by generating $50$ initial states of the robot randomly. We run an off-the-shelf motion planner on these subtasks to obtain demonstrations. We combine the solutions for all subtasks to compute critical regions according to Def. \ref{def:critical} for the given environment $E$ and goal $g$. For our experiments, we use OpenRAVE \cite{diankov10_openrave} robot simulator to simulate the motion planning problems and implementation of \emph{OMPL's} \emph{BiRRT} \cite{Berenson2009Cbirrt} by \url{https://ompl.kavrakilab.org} to compute the motion plans.

To generate the input vector, we discretize the environment into $n_d$ bins. We compute the occupancy matrix of the environment using the raster scan of the discretized environment. We append the goal value for each degree of freedom of the robot with the occupancy matrix to generate the input vector. The first channel of the label is generated by computing the criticality score for each discretized location of the environment using Def. \ref{def:critical}. We use a $3\times3$ gaussian filter to smoothen the computed regions. Additional $p$ channels are added to the label vector for each joint representing the distribution of its values for the given end-effector location.

\subsubsection{Network Training}
The layer predicting critical locations of the environment, $(CR_l)$,  uses the sigmoid activation as the task is similar to element-wise classification. The layers predicting the distribution for individual joints $(J_{l})$ use the softmax activation as the task corresponds to multi-class classification. The loss function is defined as follows:
\begin{figure*}[t]
~ \\
    \begin{center}
        \includegraphics[width=\textwidth]{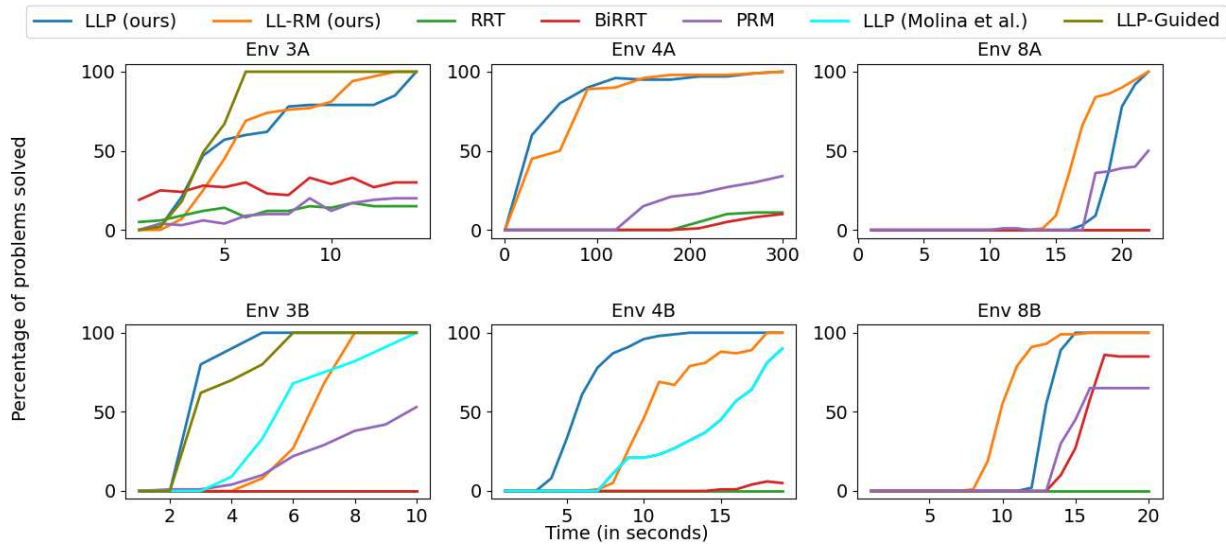}
    \end{center}
    \caption{Plots show the fraction of motion planning tasks solved (y-axis) in the given time (x-axis) for the test environments in Figures \ref{fig:test_3dof}, \ref{fig:test_4dof}, and \ref{fig:test_fetch}.}
    \label{fig:results}
    % \vspace{-1.5em}
\end{figure*}

\begin{equation}
    \mathcal{L} = \mathcal{L}_{CR} + \sum_{i=0}^{N_{\emph{dof}}} \mathcal{L}_{J}^{i}
\end{equation}
where, 
\begin{dmath*}
   %  \mathcal{L}_{CR} =  q * z * -ln(\sigma(CR_l)) + (1 - z) *  -ln(1-\sigma(CR_l)) ,
    \mathcal{L}_{CR} =  (z-1)*log(1-\sigma(CR_l)) - qzlog(\sigma(CR_l)) ,
\end{dmath*}
Here $\mathcal{L}_{CR}$ defines the weighted log loss for layers predicting the critical regions in the environment. It is important to use weighted log loss to tackle the class imbalance in the labels as most of the locations in the environment are not critical. $\sigma$ represents the sigmoid function. $q$ is the weight given to the positive samples. Also,
\begin{dmath*}
   \mathcal{L}_{J}^{i}  = - log \left(  \frac{e^{z_{k}^{i}}}{\sum_{m}e^{i}} \right)
\end{dmath*}
$\mathcal{L}_j^i$ represents the softmax cross entropy loss for $i^{th}$ joint of the robot. We use \emph{ADAM Optimizer} \citep{kingma2014adam} with learning rate $10^{-4}$. We implement the architecture shown in Fig. \ref{fig:nn_arch} using Tensorflow \cite{abadi2016tensorflow}. We train the network for $50,000$ epochs.

 \subsection{Motion Planning}
 While the biased distribution generated by our model can be used with any of the existing sampling-based motion planners, we use \emph{Learn and Link planner}~\citep{dan_llp} for our experiments. This planner generates multiple exploration trees rooted at multiple samples from the environment. It then tries to extend all the trees until they connect and form a single tree. Once a single tree is formed, the planner uses \emph{Dijkstra's Algorithm} \cite{dijkstra1959note} to compute a path from the initial state to the goal state. For our experiments, we generate $\alpha N$ samples using the learned biased distribution and $(1-\alpha)N$ samples uniformly. The planner provides two modes: $1)$ \emph{LLP}: In this mode, the exploration tree is generated on the fly. $2)$ \emph{LL-RM}: This mode is similar to \emph{PRM} where the exploration graph is built initially and reused for every problem. In our evaluations, we compare results for both modes of the planner. For our experiments, we set $N=500$ and $\alpha=0.25$.

\section{EMPIRICAL EVALUATION}
\label{sec:evaluation}

We evaluate our approach with three different types of motion planning problems. In the first set of problems, we attempt to solve path planning tasks for a rectangular robot in $\mathcal{S}\mathcal{E}^{2}$ environment. For the second set of problems, we update the rectangular robot to have a hinge joint $\omega$ and solve path planning problems for the updated robot. For the last set of problems, we use a mobile manipulator called Fetch \cite{wise16_fetch} to solve arm manipulation tasks in a cluttered environment. We compare our approach against state-of-the-art motion planners such as \emph{RRT}~\citep{lavalle1998rapidly}, \emph{PRM}~\citep{kavraki1996probabilistic}, and \emph{BiRRT}~\citep{karman2011rrtstar}. Both, \emph{PRM} and \emph{LL-RM} were given $1.0$ seconds to build the roadmap. As the \emph{Learn and Link Planner} is implemented using \emph{Python}, we use the \emph{Python} implementation of the above-stated algorithms available at \url{https://ompl.kavrakilab.org/} to compare. Our training data, code, leaned model, and results are available at \url{https://aair-lab.github.io/hdof_ll.html}.

\subsection{$\mathcal{S}\mathcal{E}^2$ domain}
\label{subsec:se3}

In this domain, the objective is to solve path planning tasks for a rectangular robot involving three degrees of freedom $(x,y,\theta)$. We use a total of $20$ environments to generate the demonstrations to train the model. Some of these environments with training labels are shown in Fig. \ref{fig:trining_envs_3}. We generate the training data as explained in section \ref{subsec:data} using $n_d=224$ and $p=10$. The input to the model is a \emph{3D tensor} of size $224\times224\times4$. Label of each input is also a \emph{3D tensor} of size $224\times224\times11$. We augment the generated data by rotating it by $90^{\circ}, 180^{\circ}$, and $270^{\circ}$. In total, we obtain $8000$ training samples from $20$ training environments.

\begin{figure}[t]
 \vspace{0.3em}
    \setkeys{Gin}{width=\linewidth}
    \begin{tabularx}{\columnwidth}{X|X}
        \includegraphics{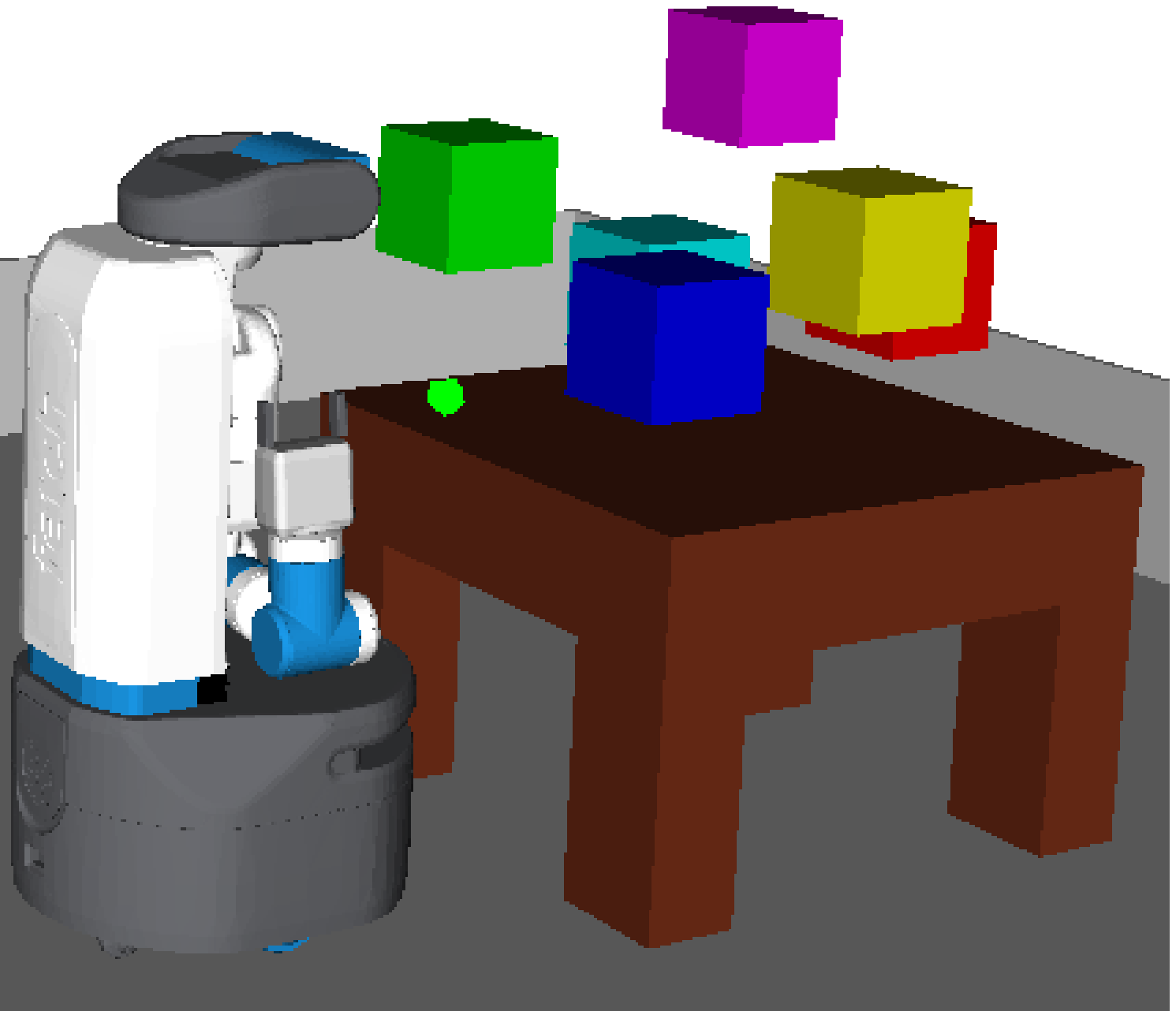} &  \includegraphics{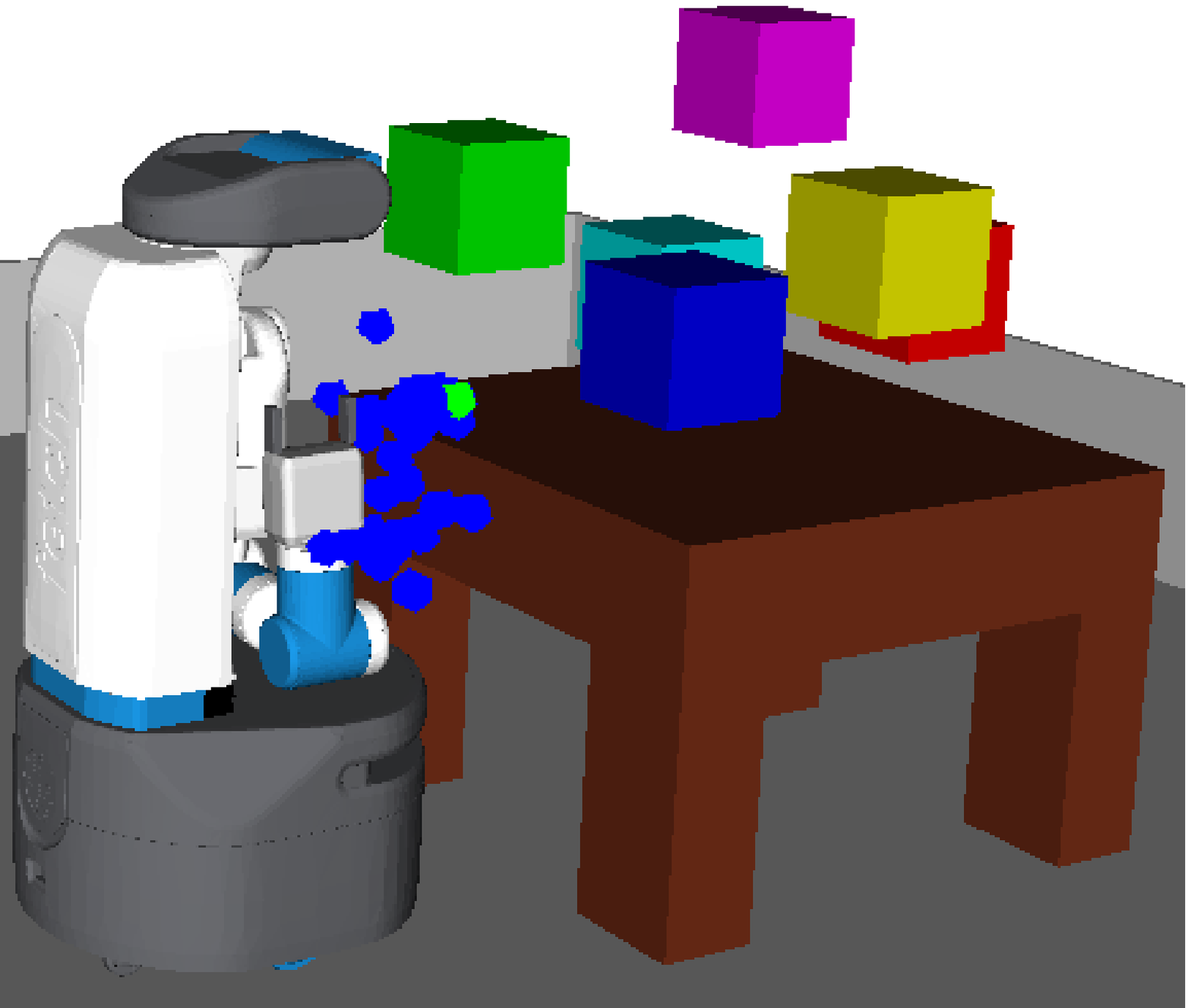}   \\
        \includegraphics{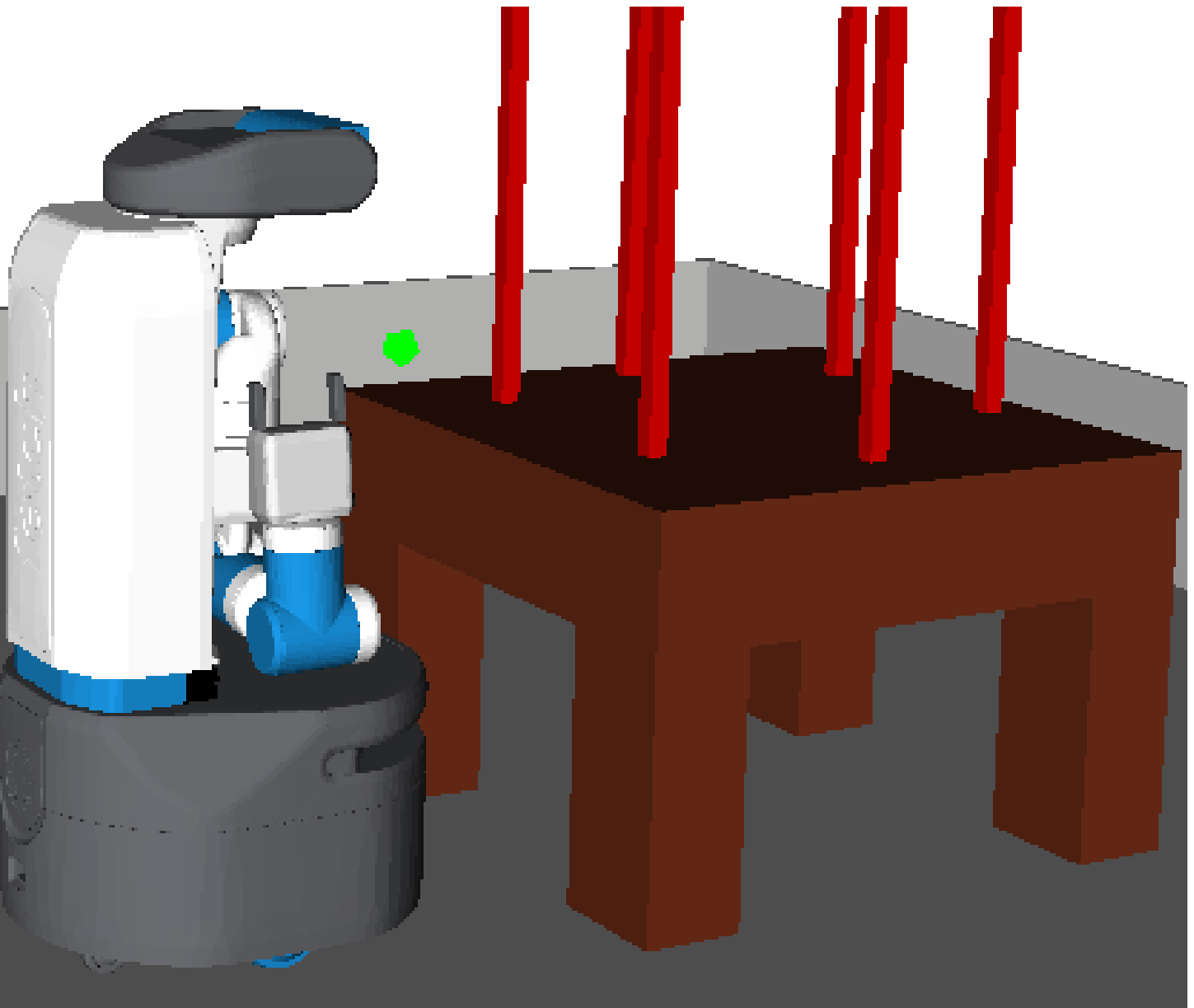} &    \includegraphics{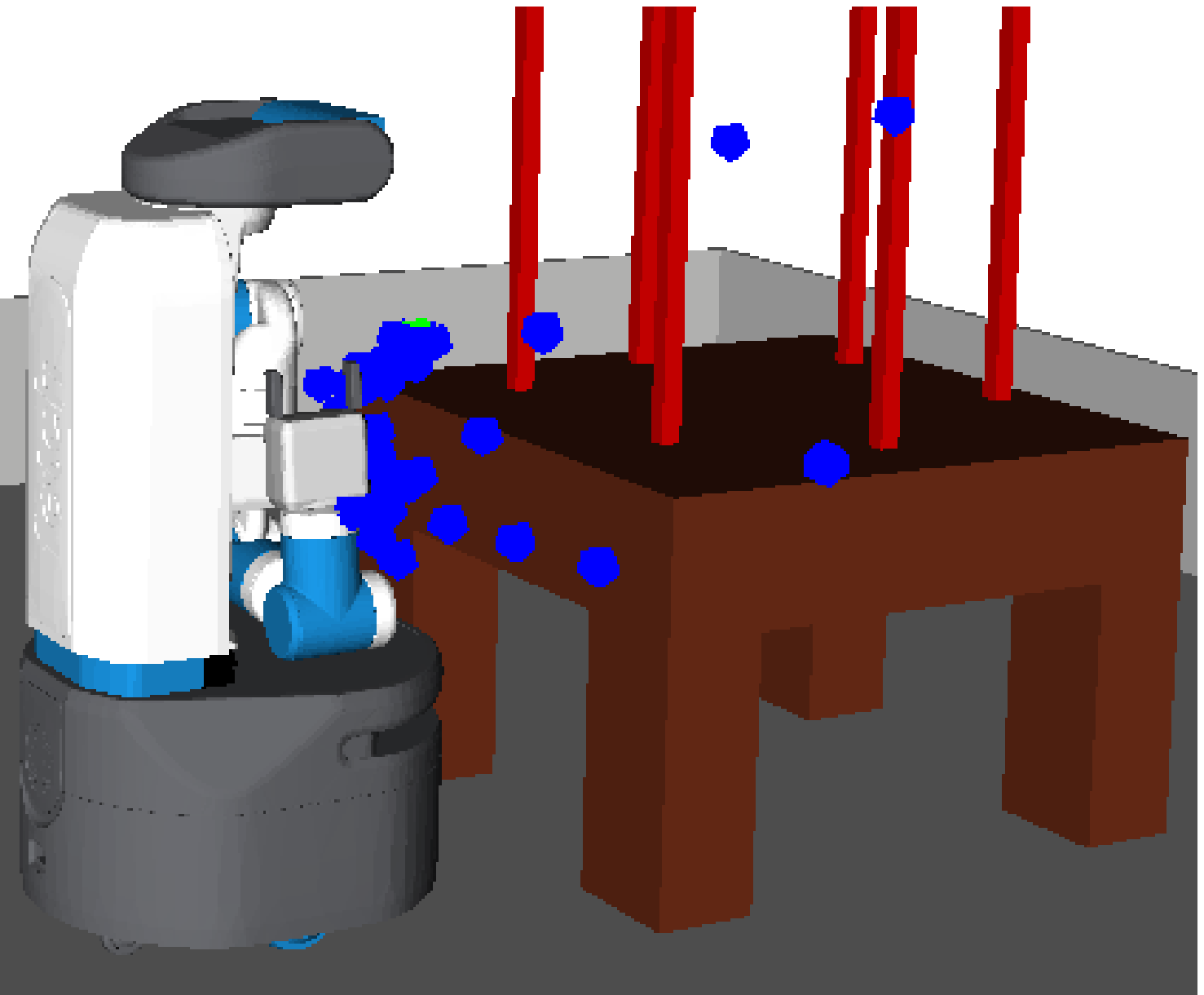}\\
        \centering (a) & \centering (b)  \\
    \end{tabularx}
    \caption{Training Environment for arm manipulation problems with Fetch. The green area shows the goal location for the end effector. Blue regions in (b) shows the critical regions for the end effector.}
    \label{fig:fetch_train}
    % \vspace{-1.5em}
\end{figure}

\subsubsection*{Result Analysis}
Fig. \ref{fig:test_3dof} shows the test environments and sampling distributions generated using our trained model for the current domain. Fig. \ref{fig:test_3dof}(a) shows the environments, Env A and Env B, with the initial state (blue) and the goal state (green) which were a part of the input to the model and Fig. \ref{fig:test_3dof}(b) and (c) shows how the network is able to learn the geometry of the environment and the robot to accurately predict the critical regions for the given problem. It can be seen from Fig. \ref{fig:test_3dof} that the model was able to identify the critical areas such as narrow passages and tunnels in the environment. It is also clear from Fig. \ref{fig:test_3dof}(c) that the model was able to learn the geometrical constraints. The model has predicted horizontal orientation of the robot as needed in narrow tunnels in Env A (top) and vertical orientation for passages for Env B (bottom).

% Fig. \ref{fig:results}(a) shows the comparison of the fraction of motion planning problems in the environments shown in Fig \ref{fig:test_3dof} using our approach and some the recent motion planning approaches. While our approach was able to solve significant number of problems in Environment A and Environment B within \textcolor{red}{XX} and \textcolor{red}{YY} using seconds \emph{LLP} mode and \textcolor{red}{XX} and \textcolor{red}{YY} using seconds \emph{LL-RM} mode respectively, the closest available approaches BiRRT and PRM solve \textcolor{red}{XX}  and \textcolor{red}{YY} problems in \textcolor{red}{XX} and \textcolor{red}{YY} seconds respectively.  

\subsection{Hinged Robot}
\label{subsec:hinged}

In this domain, the objective is to solve path planning problems for a hinged robot that has two rectangular links connected by a rotational joint $\omega$. The robot has four degrees of freedom ($x,y,\theta,\omega)$. The joint $\omega$ can rotate in the range $[-\pi/2,\pi/2]$. Similar to section \ref{subsec:se3}, we use a total of $20$ environments to generate the demonstrations to train the model. Fig. \ref{fig:training_envs_4} shows some of the environments used to generate the training data. We generate training data according to section \ref{subsec:data} using $n_d=224$ and $p=10$. The input is a \emph{3D tesnor} of size $224\times224\times5$. The label is also a \emph{3D tensor} of shape $224\times224\times21$.  We augment the environments by $90^{\circ}, 180^{\circ}$, and $270^{\circ}$ to generate the variants of the input environments. We have a total of $8000$ demonstrations including the augmented data. 

\begin{figure}[t]
 \vspace{0.3em}
    \setkeys{Gin}{width=\linewidth}
    \begin{tabularx}{\columnwidth}{X|X}
        \includegraphics{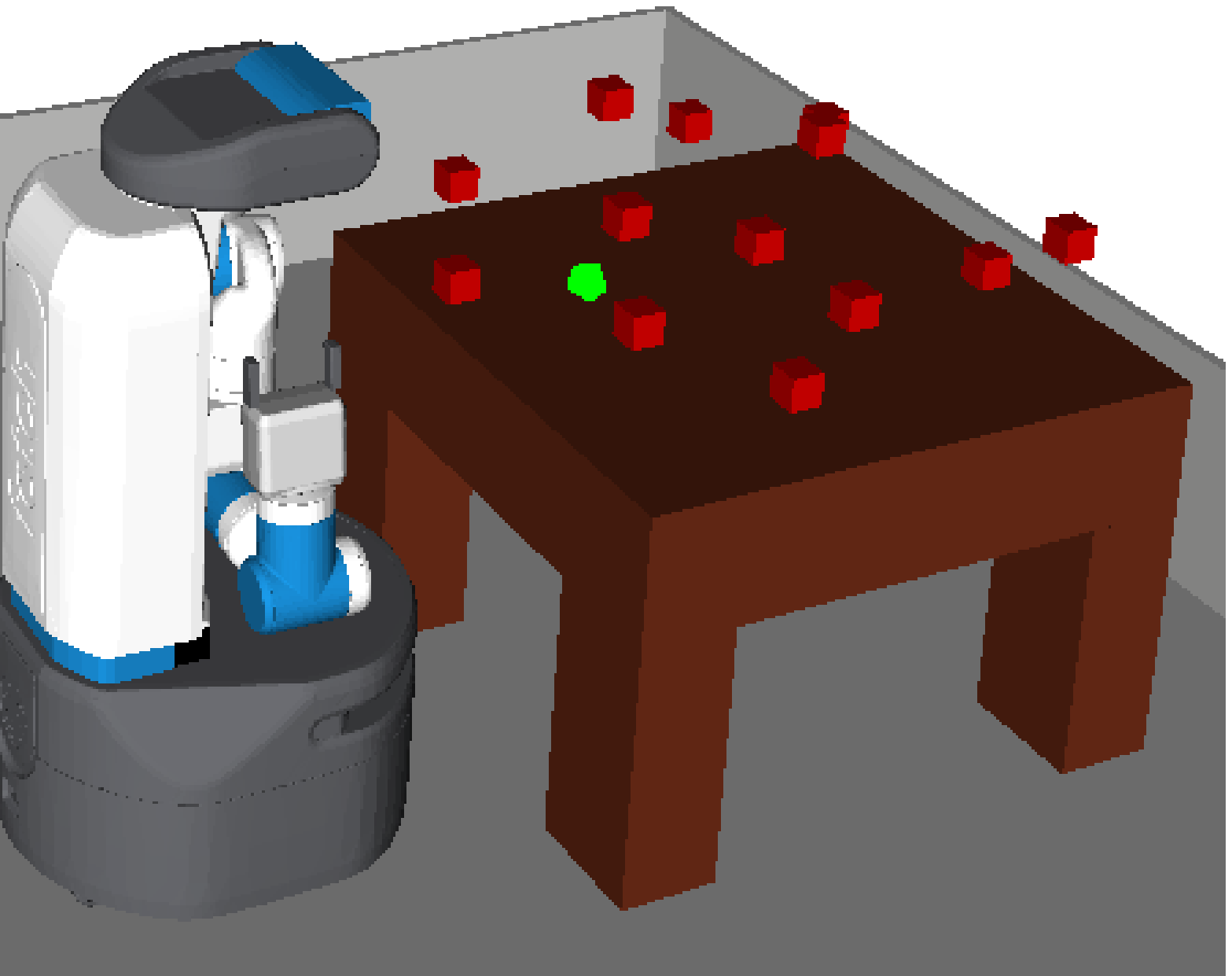} &  \includegraphics{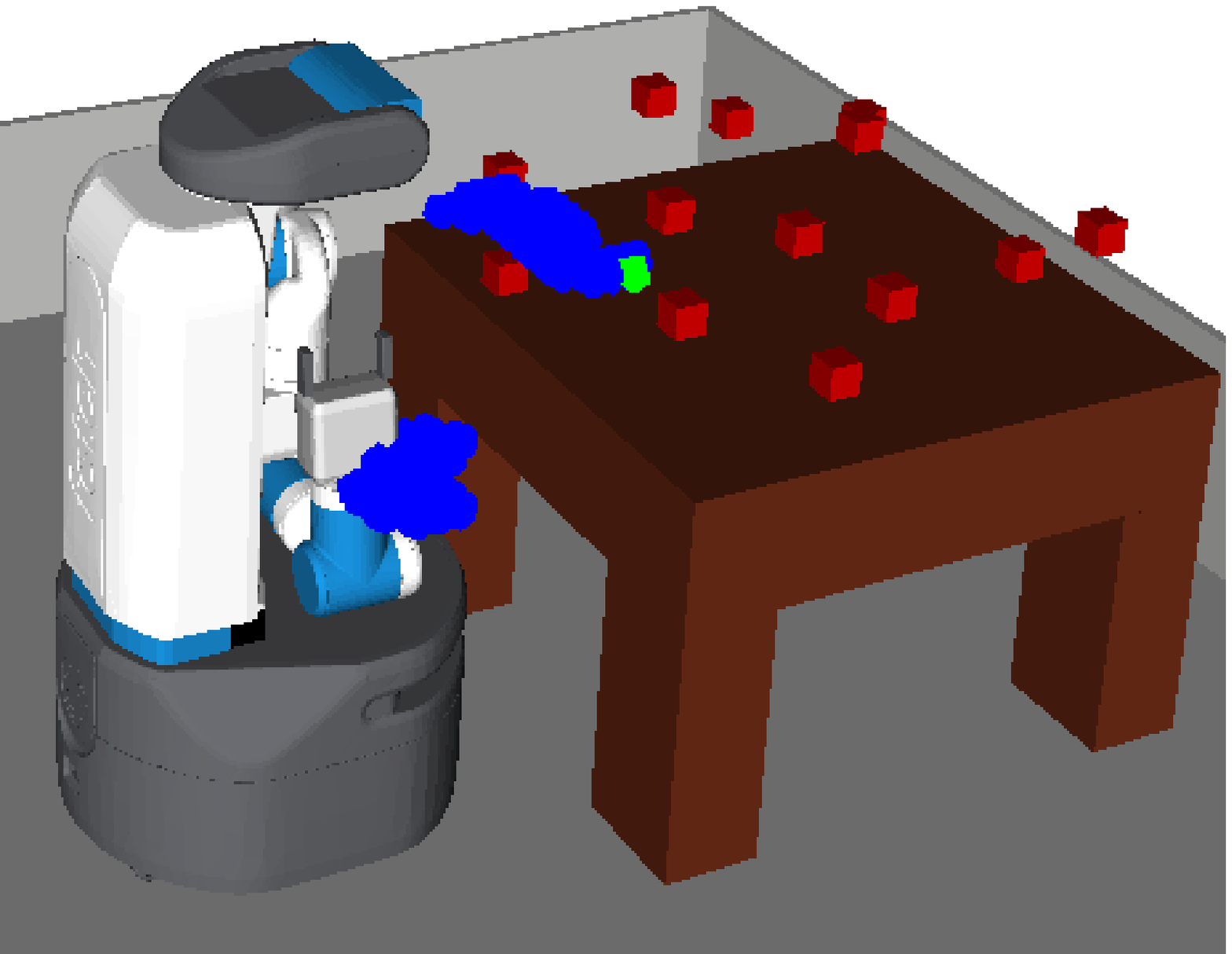}   \\
        \includegraphics{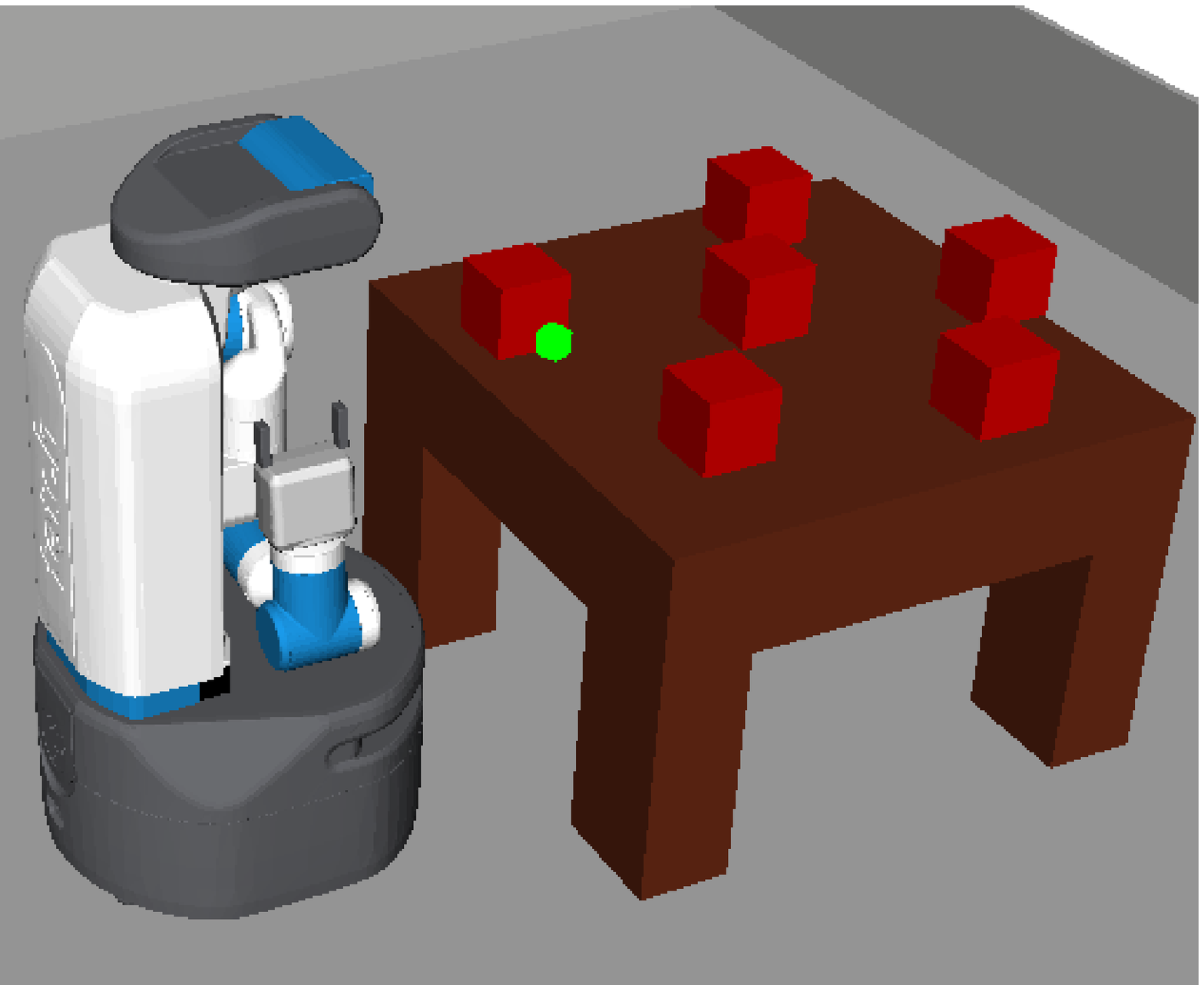} &    \includegraphics{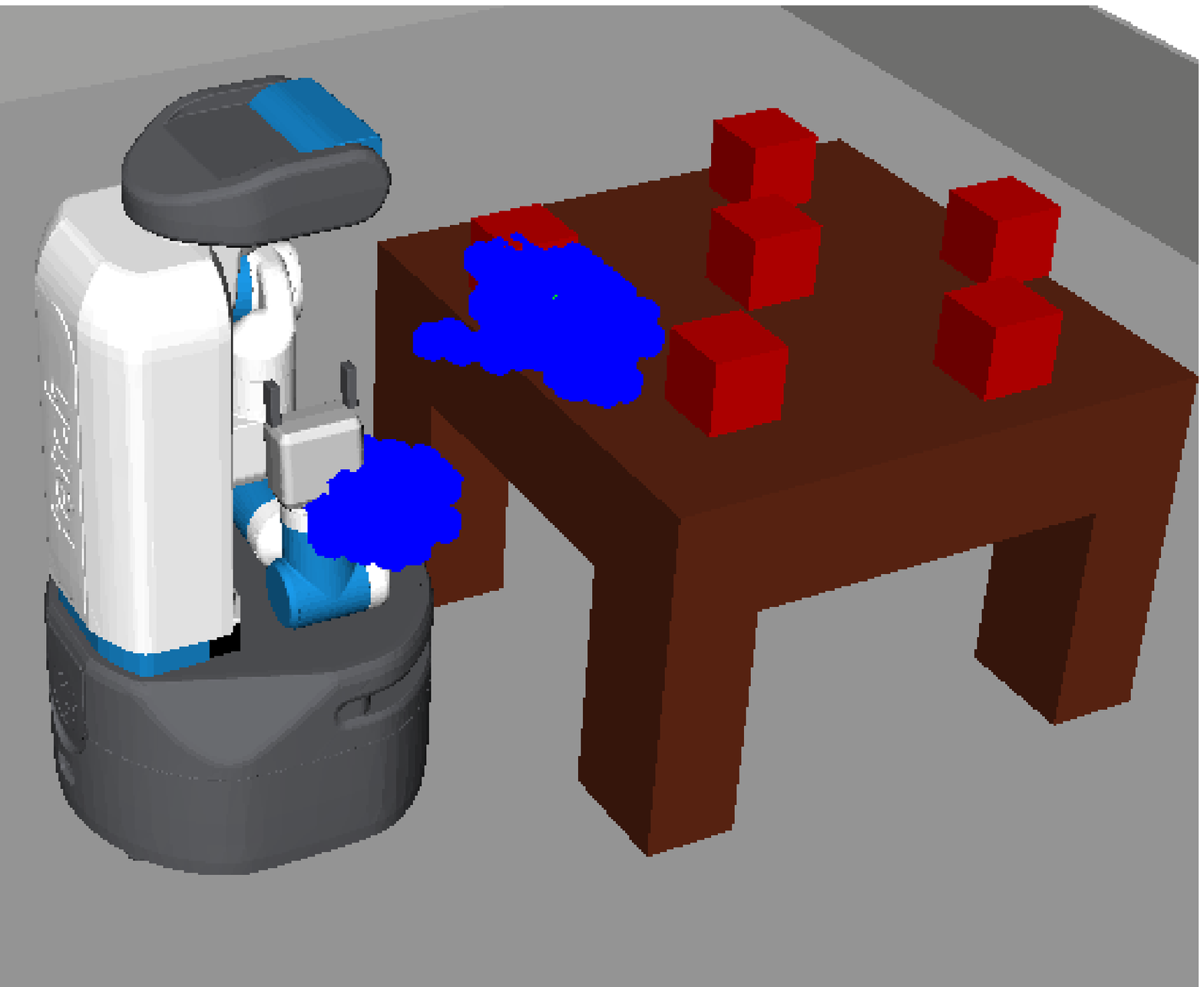}\\
        \centering (a) & \centering (b)  \\
    \end{tabularx}
    \caption{Test evaluations for arm manipulation tasks using Fetch. Top: Environment 8A and bottom Environment 8B. The green region shows the goal location for the end effector. Although the network predicts CRs for all the joints, only CRs for end-effector are shown in (b) in blue for clarity.}
    \label{fig:test_fetch}
    % \vspace{-1em}
\end{figure}

\subsubsection*{Result Analysis}
Fig. \ref{fig:test_4dof} shows the test environments and sampling distributions generated using our trained model for the \emph{hinged robot}. Fig \ref{fig:test_4dof}(a) shows the environments Env A and Env B, with the goal state (green) which was given to the model as input. Similar to the previous results, it can be seen that the model was able to learn the required distribution of critical regions as well as the geometrical constraints of the robot and the environment. Red regions in Fig. \ref{fig:test_4dof} accurately capture the critical locations in the environment. Fig. \ref{fig:test_4dof}(c) and (d) shows that the model was able to learn geometrical constraints for the problem by predicting horizontal orientation for the robot (blue regions) for narrow horizontal tunnels in Env $A$ (top) and vertical orientation for the robot (green regions) for the narrow vertical passages in Env $B$ (bottom). It is clear from Fig.~\ref{fig:test_4dof}(d) that the model was successfully able to generalize to able predict an accurate distribution for joint $\omega$. Fig.~\ref{fig:test_4dof}(d) shows that the model was able to predict a \emph{line shape} for the narrow channels and an \emph{L shape} for the corners. 

% Fig \ref{fig:results}(b) shows the comparison of the fraction of motion planning problems solved in the environments shown in Fig. \ref{fig:test_4dof} by our approach and some the recent motion planning approaches.While our approach was able to solve significant number of problems in Environment A and Environment B within \textcolor{red}{XX} and \textcolor{red}{YY} using seconds \emph{LLP} mode and \textcolor{red}{XX} and \textcolor{red}{YY} using seconds \emph{LL-RM} mode respectively, the closest available approaches BiRRT and PRM solve \textcolor{red}{XX}  and \textcolor{red}{YY} problems in \textcolor{red}{XX} and \textcolor{red}{YY} seconds respectively.  

\subsection{Arm Manipulation: Fetch robot}

For the last set of problems, we use a mobile manipulator named \emph{Fetch}~\citep{wise16_fetch} with \emph{8 DOFs} to perform arm manipulation. The goal of these experiments is to show that our approach is not limited to the path planning tasks and is generalizable to any arbitrary motion planning task. Fig. \ref{fig:fetch_train} shows two of the eight training environments used to generate demonstrations. We generate a total of $800$ training samples according to section \ref{subsec:data} for eight training environments using $n_d=64$ and $p=10$. While the previous experiments involved robot movements along only $x$ and $y$ axis, these experiments included arm movements along all three $x$, $y$, and $z$ axis. We update the architecture shown in Fig. \ref{fig:nn_arch} by replacing convolutional and deconvolutional layers with \emph{3D} convolutional and \emph{3D} deconvolutional layers. We also reduce the number of discretized locations in the environment as the reach of the robot is much lesser for arm manipulation compared to path planning problems. The input to the model is a \emph{4D tensor} of size $64 \time 64 \times 64 \times 11$. The first channel consists of the distribution for critical locations in the environment. Label for each input is also a \emph{4D tensor} of size $64\times64\times64\times81$.

\subsubsection*{Result Analysis}

Fig. \ref{fig:test_fetch} shows the test environments and sampling distributions predicted by our model. Fig. \ref{fig:test_fetch}(a) shows the input environment to the model. Fig. \ref{fig:test_fetch}(b) shows the critical locations predicted by the model. It can be seen from Fig. \ref{fig:test_fetch}(b) that the model was able to predict critical locations around the goal. For simplicity, we do not show distributions for each joint, but our results show that the model was able to learn these distributions and improve the efficiency of motion planning.

\subsection{Comparison}

Fig. \ref{fig:results} shows the comparison of the number of motion planning problems solved by our approach in a given time with the available SBMPs. The \emph{x-axis} shows the time (in seconds) and the \emph{y-axis} shows the fraction of motion planning problems solved (out of $100$) in the given time. While sampling-based motion planners such as \emph{RRT}, \emph{BiRRT}, and \emph{PRM} struggle to compute motion plans due to the inability to sample critical configurations of the environment, our biased sampling distribution helps the \emph{Learn and Link planner} to efficiently sample such configurations by using the learned model. It can be seen from Fig. \ref{fig:results} that our approach was able to solve a significantly higher number of problems in less than $20\%$ time than the available sampling-based motion planners with uniform samplers. We also compare our approach in the domains \ref{subsec:se3} and \ref{subsec:hinged} with the approach developed by \citet{dan_llp}. While their approach only predicts critical regions for \emph{2D spaces}, we uniformly sample remaining joint values for the robot. Results in Fig. \ref{fig:results} for environments  \emph{3A} and \emph{3B} show that the approach developed by \citet{dan_llp} was able to perform better than the current sampling-based approaches like \emph{RRT} and \emph{BiRRT}, but our approach was able to outperform it significantly due to the availability of critical regions for all the degrees of freedom compared to just for the base location. 

We also implemented \emph{Guided-LLP} mode that uses identified critical regions to formulate a high-level search problem. We convert the critical regions to high-level states and construct a high-level search space. For the basic implementation, we use the minimum distance between two regions to compute the connectivity between high-level states. For our experiments, we set this distance to $0.2m$. For the given initial and goal state, a high-level plan is computed by searching the shortest path in the high-level search space that is then used to direct the low-level \emph{LLP}. We evaluate this mode in environments \emph{3A} and \emph{3B}. Fig. \ref{fig:results} shows that for environment \emph{3A},  \emph{Guided-LLP} was able to perform slightly better compared basic variant. For environment \emph{3B}, its performance was slightly inferior compared to the basic version. This performance can be improved by using a more intelligent approach for establishing connectivity between high-level states.

\section{ABLATION STUDY}
\label{sec:ablation}

Apart from the presented approach that uses UNet architecture shown in Fig. \ref{fig:nn_arch}, we mainly tried two other approaches. As part of the first approaches, we implemented a generative network known as \emph{conditional generative adversarial network} (CGAN) \cite{mirza2014cgan} to predict the criticality score for all the states in an environment. The approach was not able to scale due to the dimensionality of the model. For the second approach, we tried to learn the distribution of each joint independently without identifying critical locations for the end-effector (base link for path planning) of the environment. While the approach was able to learn distributions, it failed to improve the efficiency of the planner due to the correlation between the values of each joint.

\section{CONCLUSION}

In this paper, we present an approach to learn biased sampling distribution that works with any available sampling-based motion planner for any arbitrary motion planning problem. Our model uses demonstrations to learn critical states of the given environment and sampling distribution for each joint of the robot. Through our experiments, we show that our approach is robust and can be scaled to robots having higher degrees of freedom.

\section*{Acknowledgements}
This work was supported in parts by the NSF under grants
IIS 1844325, IIS 1909370, and OIA 1936997.

% \newpage

% \bibliographystyle{aaai21}
\bibliography{ref.bib}

\end{document}